\documentclass[manuscript]{acmart}
\usepackage{geometry}% http://ctan.org/pkg/geometry
\usepackage{graphicx}% http://ctan.org/pkg/graphicx
\usepackage{xspace}
\usepackage{soul}%highlight
\usepackage{multirow}

\usepackage{algorithm} % algorithm
\usepackage{algorithmic}
\usepackage{enumitem} % for left margin
\usepackage{url}
\usepackage{hyperref}
\usepackage{balance}
\usepackage{siunitx}
\usepackage{subcaption}
\newcommand{\etal}{\emph{et al.}\xspace}
\newcommand{\ie}{\emph{i.e.,}\xspace}
\newcommand{\eg}{\emph{e.g.,}\xspace}
\newcommand{\vs}{\emph{v.s.}\xspace}
\newcommand{\etc}{\emph{etc.}\xspace}

\newcommand{\noind}[0]{\noindent}
\newcommand{\noindpar}[1]{\noind {\bf #1}}

\AtBeginDocument{%
  \providecommand\BibTeX{{%
    \normalfont B\kern-0.5em{\scshape i\kern-0.25em b}\kern-0.8em\TeX}}}

\copyrightyear{2025}
\acmYear{2025}
\setcopyright{rightsretained}
\acmDOI{00.0000/0000000.0000000}
\acmISBN{000-0-0000-0000-0/00/00}
    
\begin{document}
\abovedisplayskip=1pt
\abovedisplayshortskip=0pt
\belowdisplayskip=2pt
\belowdisplayshortskip=0pt

\title[Arm Robot: AR-Enhanced Embodied Control and Visualization for Intuitive Robot Arm Manipulation]{Arm Robot: AR-Enhanced Embodied Control and Visualization for Intuitive Robot Arm Manipulation}
% freeze, mirror mode, scaling, digital twin, buffer.

% \title[Hand Robots: Correspondence Control in AR for Free-Hand Robot Programming by Demonstration]{Hand Robots: Correspondence Control and Editing Techniques in AR for Free-Hand Robot Programming by Demonstration}
% \title{Using AR based end-effector embodiment to enable light-weight user-centered observation-based robot training by demonstration}


\def\systemname {\textit{Arm Robots}\xspace}

\author{Siyou Pei}
\affiliation{%
  \institution{University of California, Los Angeles}
  \city{Los Angeles}
  \state{California}
  \country{USA}}
\email{sypei@ucla.edu}

\author{Alexander Chen}
\affiliation{%
  \institution{University of California, Los Angeles}
  \city{Los Angeles}
  \state{California}
  \country{USA}}
\email{a.chen711@ucla.edu}

\author{Ronak Kaoshik}
\affiliation{%
  \institution{University of California, Los Angeles}
  \city{Los Angeles}
  \state{California}
  \country{USA}}
\email{ronak42@g.ucla.edu}

\author{Ruofei Du}
\affiliation{%
  \institution{Google Research}
  \city{San Francisco}
  \state{California}
  \country{USA}}
\email{me@duruofei.com}

\author{Yang Zhang}
\affiliation{%
  \institution{University of California, Los Angeles}
  \city{Los Angeles}
  \state{California}
  \country{USA}}
\email{yangzhang@ucla.edu}

\renewcommand{\shortauthors}{Pei et al.}

\begin{abstract}
%version 4
Embodied interaction has been introduced to human-robot interaction (HRI) as a type of teleoperation, in which users control robot arms with bodily action via handheld controllers or haptic gloves. Embodied teleoperation has made robot control intuitive to non-technical users, but differences between humans' and robots' capabilities \eg ranges of motion and response time, remain challenging. In response, we present Arm Robot, an embodied robot arm teleoperation system that helps users tackle human-robot discrepancies. Specifically, Arm Robot (1) includes AR visualization as real-time feedback on temporal and spatial discrepancies, and (2) allows users to change observing perspectives and expand action space.
We conducted a user study (N=18) to investigate the usability of the Arm Robot and learn how users perceive the embodiment. Our results show users could use Arm Robot's features to effectively control the robot arm, providing insights for continued work in embodied HRI.

\end{abstract}

\begin{teaserfigure}
  \centering
  \includegraphics[width=\linewidth]{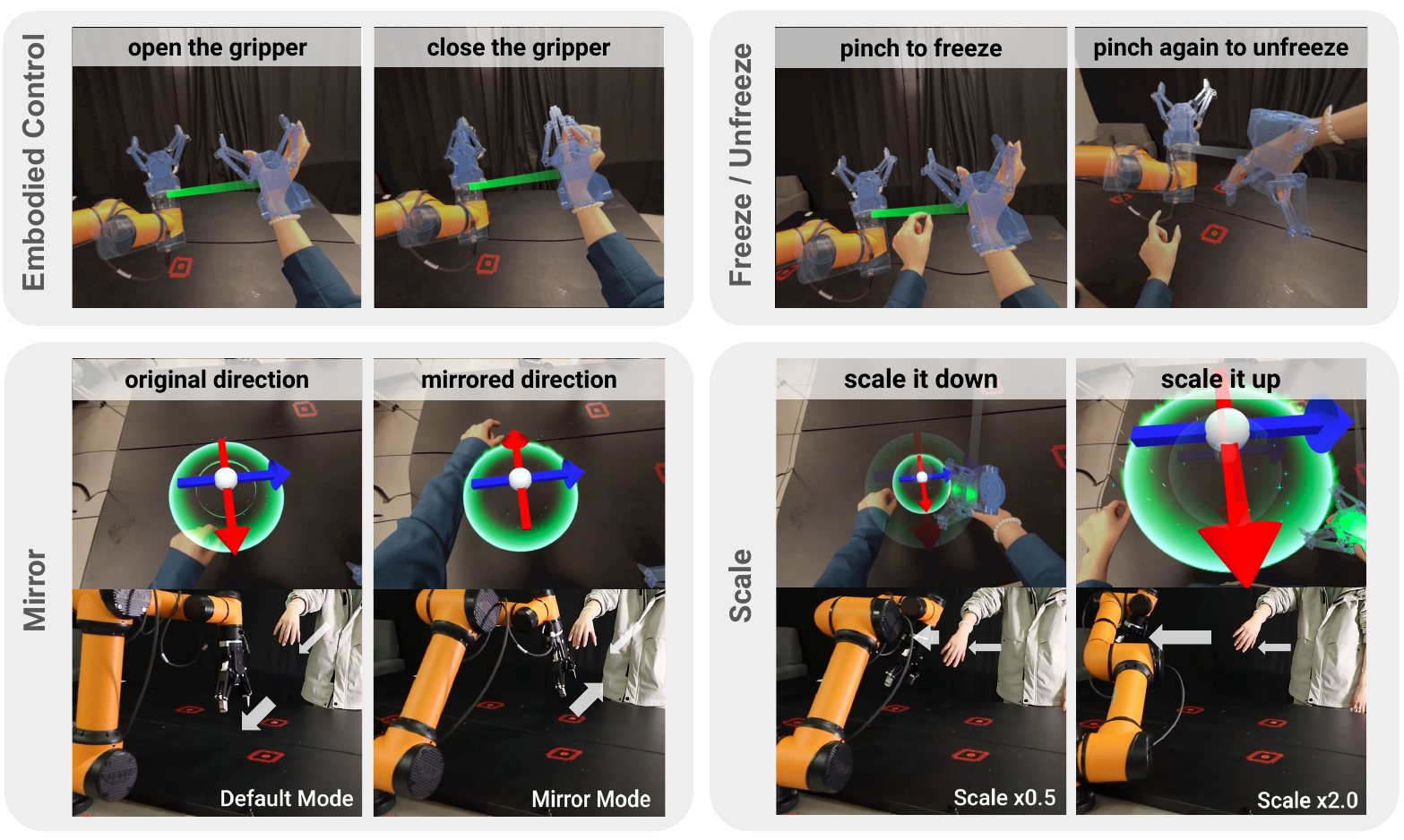}
    % cannot add an empty line between this line and next line, otherwise figures cannot render
  \caption{Arm Robot is a set of AR-enhanced embodied interaction techniques with visual feedback for intuitive robot manipulation. The robot's gripper closes as the user's hand closes (upper left). The ``Freeze/Unfreeze'' feature (upper right) allows users to pause or resume embodiment. With ``Mirror'' (lower left) mode, users can intuitively map their motion to the robot's motion in a mirror-like manner. The ``Scale'' function (lower right) enables users to adjust the scale of embodied movement. The coordination disk (green circle with two arrows) affords interactions for ``Scale'' and ``Mirror'', while a pinchable line (the green/grey transparent line between hand and physical gripper) supports ``Freeze/Unfreeze''. The line's color indicates control ON (green) or OFF (grey). A virtual robot, superposed on the physical robot, embodies the real-time response of the robot to user interaction.}
  \Description{This teaser figure consists of a series of photographs labeled, which demonstrate different functions of the AR system. In embodied control, a person's hand is shown wearing a device that controls a robotic gripper, depicted with augmented reality overlays. The robotic gripper closes as the user's hand closes, showcased in a progression from open to closed positions. The "Freeze/Unfreeze" feature allows users to pause or resume the control of the robotic hand. A virtual line appears in the user's view, changing color from green to grey to indicate whether the control is active or paused. The "Scale" function, enables users to adjust the magnitude of their movements that are translated to the robot. A coordination disk with colored axes is presented, changing in size to represent the scaling effect. In "Mirror" mode, users can intuitively map their motion to the robot's movements in a mirror-like manner. The coordination disk is also present in these images, indicating the mirroring interaction. A virtual representation of the robot's gripper is superimposed on the physical robotic hand, showing the real-time response of the robot to the user's hand movements.}
  \label{fig:teaser}
\end{teaserfigure}

\maketitle

\section{Introduction}
\label{intro}

Using hands to intuitively control robot arms has been demonstrated in human-robot interaction (HRI), where users could control a robot arm with bodily movements via hand instrumentation such as controllers \cite{su_mixed-reality-enhanced_2022} or gloves \cite{fritsche_first-person_2015}. Unlike conventional robot programming methods, these embodied interaction methods made robot arms readily available for new tasks that emerged in-situ, and easy to use for non-technical users. However, there remain usability challenges caused by spatial and temporal discrepancies between humans' and robots' capabilities. 

Regarding \textit{spatial discrepancies}, humans and robots have different numbers of sections, arm lengths, and joint kinematics, which leads to different ranges of motion. For example, the motion range of a warehouse robot can be much bigger than a human arm reach, while a surgical robot may have finer ranges of motion. In addition, while the robot gripper follows the human hand, the rest of the robot arm usually has a different pose than the human arm, which may appear between users and the gripper, blocking observation of the gripper's status. Furthermore, some gripper orientations might require hand poses that are difficult to perform. For example, when the user and the robot arm are on opposite sides of the task, any angles of the robot gripper pointing towards the user would require the user to bend their hands awkwardly towards themselves. 

For \textit{temporal discrepancies}, humans and robots, if thought of as systems, have different response times. Specifically, most commercial robot arms have a non-negligible latency referring to human movements as ground truth benchmarks. System latency comes from mechanical maximum acceleration, network, control algorithms \etc. For example, human body itself is a highly delicate system that can move a hand at up to \SI{45}{m/s} (\SI{100}{mph}) within 20 to 40 millisecond \cite{schwartz2016movement}, while a robot arm's maximum speed is usually 1-2 m/s with a peak acceleration of \SI{4}{m/s^{2}} \cite{universal_robot_collaborative,franka_robotics_franka}. Also, communication of tracking data and robot commands worsens the latency. Prior work has shown that the gap in response time increases the cognitive load of robot control \cite{lum_teleoperation_2009, blackett_effects_2022}. 

These discrepancies severely limit the usability of current embodied interaction techniques and motivated this work to identify additional interaction techniques to make embodied interaction in HRI even more powerful. To help users tackle human-robot differences in capabilities, we present Arm Robot, an embodied robot arm control system that allows users to change the human-robot spatial mapping to accommodate differences in ranges of motion. With Arm Robot, a user can adjust the spatial mapping between human and robot, which includes the on/off (``Freeze / Unfreeze''), the scale (``Scale''), and the mirror reversion (``Mirror'') of embodied movement. Additionally, Arm Robot accommodates users with zero-delay robot visualization as real-time feedback. 

%We need an interaction sequence to explain this!
As the robot and the operator are in the same room in our settings, we naturally turn to AR for in-situ interaction and visualization. To use Arm Robot, a user could pause/resume the control with ``Freeze / Unfreeze'', scale up or down the spatial mapping with ``Scale'', or reverse motion on one axis using ``Mirror''. When their range of motion is not compatible with robot's, they can flexibly combine these techniques to extend their movements, and tackle awkward hand pose. For example, by using the ``Freeze / Unfreeze'' feature, users are allowed to freeze the robot at their limit, relocate and reposition, unfreeze, and continue their task. When the robot cannot move as fast as instructed, they could preview robot action by referring to the zero-delay robot visualization. We elaborate on how each feature/visualization was designed to augment users' capabilities or perception in Sec \ref{design}. This design of Arm Robot concludes our first contribution -- proposing an embodied robot control system that augments human perception and capability to compensate for the human-robot discrepancies. These interaction designs were solicited by literature, refined after a pilot study, and proven effective in a later main user study.

Besides addressing the limitations of embodied methods in robot arm control, we intend to fill a vacuum in HRI studies of embodiment as few studies have investigated the embodiment in HRI. This compares drastically different than much investigations that have been done in XR research \cite{pei_hand_2022, cordeil_embodied_2020, feuchtner_extending_2017,abtahi_beyond_2022}, which even include a specific thread of investigations of freehand interaction \vs controller-based interaction \cite{rantamaa2023comparison, khundam2021comparative}. We conducted a user study (N=18) to investigate how the adjustable spatial mapping and visualization impact usability overall, and the sense of embodiment which has been a key to the ease of use of embodied interactions. In addition, we considered both freehand and controller-based embodied interactions in our study for a larger front where new knowledge could be generated by comparisons of these two that feature different extents of embodiment -- freehand has higher extends than controller-based method. We designed a variety of robot arm tasks and collected both quantitative and qualitative data in our study. 

Results show Arm Robot helped participants address the the HCI challenges in robot teleoperation. 
The usefulness of methods to adjust spatial mapping ranked as follows, from most useful to least useful: ``Freeze / Unfreeze'', ``Scale'', and ``Mirror''.
Participants also appreciated visualizations in augmenting their perception of discrepancies. 
Furthermore, participants held various preferences on the extent of embodiment. Our findings provide insights for future work in improving robot teleoperation experience. 

In summary, we focus on the following research questions:
\begin{itemize}[nosep, leftmargin=*]
    % \item \textbf{RQ1}: What is embodied interaction in XR?
    \item \textbf{RQ1}: What are the challenges of embodied interactions in robot arm control?
    % \item \textbf{RQ2}: Does the digital twin visualization improve user experience and why? (digital twin)
    \item \textbf{RQ2}: How do we design a system to address these challenges in embodied robot arm control?
    \item \textbf{RQ3}: How do freehand and controller-based Arm Robot affect the usability differently?
\end{itemize}

In response, our contribution unfolds in three aspects,
\begin{itemize}[nosep, leftmargin=*]
    % \item Proposing the theory of a continuous embodiment spectrum in XR and a design space of embodied control for robot arm manipulation; 
    \item \textbf{Identify challenges}: We revealed existing HCI challenges of robot teleoperation -- difficulty in understanding motion mapping, limited feedback bandwidth, and limited action space due to discrepancies in range of motion.
    \item \textbf{System design}: In response, we designed Arm Robot, an AR-enhanced embodied control system that enabled discrepancy visualizations and adjustable spatial mapping.
    \item We evaluated the effectiveness of Arm Robot and present \textbf{findings} about the extent of body embodiment users needed for effective embodied teleoperation;
\end{itemize}
\section{Related Work} \label{related}

\subsection{XR-enhanced HRI}

XR has been used to enhance interaction with robots.
% % list the work that uses (mainly) VR for an immersive experience of a reconstructed remote environment (exists in the world), or a virtual environment (does not exist, purely virtual)
VR in HRI is often used in data collection or video streaming. For example, Zhang \etal \cite{zhang_deep_2018} used virtual scenes in VR to collect low-cost training data, while other VR applications aim to facilitate safe training or remote operation in a reconstructed real-world environment \cite{gaurav_deep_2019, togias_virtual_2021, matsas_design_2017}, or from a camera's perspective \cite{delpreto_helping_2020, yim_wfh-vr_2022,arunachalam_holo-dex_2022, xu_robot_2018,su_mixed-reality-enhanced_2022}.

% % list the work that uses AR for annotation and feedback, e.g., the target placement of an object, the trajectory in the air, color-coded status of the robot/task, a virtual robot (e.g. when people don't have a physical one)
While VR aims for immersion in remote environments, AR/MR in HRI aims to improve usability for physically co-located settings. Suzuki \etal \cite{suzuki_augmented_2022} present a comprehensive survey on AR-enhanced HRI, and summarised the variety of AR design components including UIs/widgets \cite{wang_explainable_2023}, spatial references and visualizations \cite{soares_programming_2021, wang_explainable_2023}, and embedded visual effects. Digital twins of robots are one kind of spatial visualization. Prior art has investigated how digital twins of robots and objects in AR enable users to program a robot even without a physical one \cite{bambusek_combining_2019, duan_ar2-d2training_2023, kaarlela2022digital}. Closest to our work, Wang \etal \cite{wang_explainable_2023} used a digital twin for task planning to facilitate human-robot collaboration, when a physical robot awaits to execute. We also deploy digital twin of a co-located robot. However, our digital twin is designed to tackle system latency in real-time teleoperation instead of advanced planning. The bases of virtual and physical robots always align for users to compare their poses and trajectories in real-time. 
We also leveraged AR to enable other interaction with 3D virtual interfaces (\eg the line, the disk, and the arrows in Sec \ref{design}) rather than 2D UI or widgets. 
% The design of Arm Robot can also be applied in a remote setting with a reconstructed 3D environment in VR. 

\subsection{Embodied Interaction in XR}
Embodied interaction \cite{dourish_where_2001} has a unique strength in transforming abstract computation into intuitive interactions, using metaphors of practical physical actions or social conventions \cite{antle2011embodied}. 
As XR technologies emerge, embodied interaction has been widely used in XR for natural spatial input, adopting new metaphors beyond traditional WIMP in 2D screens \cite{laviola20173d}.  
% We believe human body embodiment in XR spans across a spectrum rather than a binary classification. Many factors can affect the embodiment level in the spectrum, \eg the degree of spatial correlation, the similarity of functions, or the degree of human body engagement. We take the degree of spatial correlation as the metric to organize related work. 
Human body is a common vessel in embodied interaction. For example, Hand Interfaces \cite{pei_hand_2022} use a hand to embody a virtual joystick with a thumb-up gesture. 
Embodied Exploration \cite{pei_embodied_2023} allows wheelchair users to embody a virtual avatar in VR to facilitate intuitive and precise remote assessment of environment accessibility. Different spatial mappings between the human body and virtual counterpart have been explored separately in prior work of body-embodied interaction. For instance, Sousa \etal \cite{sousa2019warping} and Tao \etal \cite{tao_embodying_2023} introduce a natural warping of virtual avatars to increase annotation accuracy or realism. The spatial offset in mapping can also be made large on purpose, for example, Feuchtner \etal \cite{feuchtner_extending_2017} and Schjerlund \etal \cite{schjerlund_ninja_2021} intentionally depart virtual hand(s) by a large offset to augment reachability. In addition, the scale of spatial mapping was also studied in Mini-Me \cite{piumsomboon_mini-me_2018} and Snow Dome \cite{piumsomboon_snow_2018}, which change the size of embodied avatars to accommodate various collaboration scenarios.

While prior works present different spatial mappings separately, Arm Robot allows users to modify the spatial mapping with ``Freeze / Unfreeze'', ``Scale'', and ``Mirror'' features, and explores how users would take advantage of these features during embodied teleoperation.

\subsection{Embodied Interaction in HRI}
\label{sec:challenges}
Embodied interaction has been used to control robot arms, where the extent of body embodiment varies significantly. 3D Mouse \cite{radalytica_precise_2020} by Radalytica is a freedrive button to shift, tilt, and press, which passes the same actions to a robot end effector. The media of embodiment is the 3D Mouse instead of the human body. Instead, embodied teleoperation with handheld controllers involves more body embodiment by directly using 6-DOF hand location as input, employing a VR controller \cite{su_mixed-reality-enhanced_2022}, or a pen-shaped tracker \cite{nordbo_robotics_mimic_2022}. Embodied teleoperation with free hands engages the human body to the next level, as users can deliver more unrestricted motion to control the robot through hand-worn sensors \cite{fritsche_first-person_2015, li_mobile_2020}, or vision-based methods \cite{qin_anyteleop_2023, sivakumar_robotic_2022, arunachalam_holo-dex_2022, duan_ar2-d2training_2023}.

While various embodiments of teleoperation interfaces offer distinct advantages, the common thread is the need for intuitive, seamless control of robot arms, which becomes particularly challenging when the nuances of human-robot interaction are considered. Building on this diversity in control methods, our research explores the human-centered challenges encountered in real-world teleoperation scenarios, as identified through expert consultation. Specifically, through consulting with three experts who use robot arms daily and have more than one year of experience in robot teleoperation, we identified several HCI challenges in robot teleoperation in response to \textbf{RQ1} (annotated with warning signs in Fig.~\ref{fig:cycle}). In this paper, we aim to address these HCI challenges by introducing AR visual cues to maximize the utility of prior knowledge, providing real-time feedback, and enable a larger action space. We also studied how the extent of body embodiment affected the usability and sense of embodiment in teleoperation by comparing freehand Arm Robot and a controller-based Arm Robot.
%In Arm Robot, a digital twin of the robot has zero latency, superposing the physical one. As users move their hands, they can observe the virtual robot as real-time feedback, and wait for the physical robot to catch up, until they fully overlap again. 
%To help users tackle human-robot discrepancies, prior work has explored a predictive display \cite{dybvik_low-cost_2021} to account for system latency. 
In comparison with prior work (e.g., \cite{dybvik_low-cost_2021}), Arm Robot offers a more comprehensive solution to address gaps in both ranges of motion and response time.

\noindpar{Challenge\#1: Lacking an understanding of robot motion:} The experts reported the challenge of lacking an understanding of robot motion.In particular, they commented that people without robot expertise usually find it harder to perform teleoperation because they lack prior knowledge about the robot's range of motion and hand-gripper mapping. They explicitly mentioned that the difference between experts and novices in teleoperation is their understanding of how hand motion is regarded by the robot gripper. 

\noindpar{Challenge\#2: Inaccurate observation:} 
The second challenge is inaccurate observation, primarily caused by occlusion and time delay. During teleoperation, the robot arm often blocks the operator's line of sight, and users cannot change their perspective due to embodied movement. Although some systems use multiple cameras for better feedback, perceiving depth from 2D video remains difficult, requiring even experts to rely on spatial imagination. Time delays in robot systems, stemming from control algorithms and communication, further disrupt the perception-action cycle, especially with larger robots requiring trajectory smoothing. These delays vary unpredictably across systems and movements.

%The second challenge is inaccurate observation. Inaccurate observation comes from occlusion and time delay. During teleoperation, visual occlusion happens frequently because the arm portion of the robot might block the line of sight when moving. People cannot change their position to alter their observation perspective because their movement is always embodied. Some existing pipelines introduced multiple cameras to stream different perspectives on a screen for richer feedback but perceiving depth from videos is difficult. Therefore, even the experts have to count on spatial imagination. 
%On the other hand, according to the experts, time delays in robot systems are unavoidable due to low-level control algorithms and communication, especially for bigger robot arms with higher payloads that require extra trajectory smoothing. The lag of robot motion leads to the lag of feedback in the perception-action cycle. Time delay also varies across robot brands, the efficiency of algorithms, and the complexity of movement, so it is not as easy to predict. 

\noindpar{Challenge\#3: Limited action space:} Finally, the third challenge is limited action space. Because the robot arm and human arm have different ranges of motion, humans cannot reach some positions because of arm length and joint flexibility (See Video Figure at 1:46 and 2:40 for examples). Therefore, they need to find a perfect angle that their hand motion feels natural through trial and error when recording a teleoperated demonstration.

\section{Interaction Design}
\label{design}
\subsection{Iteration 1: Interaction Ideation through Low-Fidelity Prototyping}
\begin{figure}
    \centering
    \includegraphics[width=0.5\linewidth]{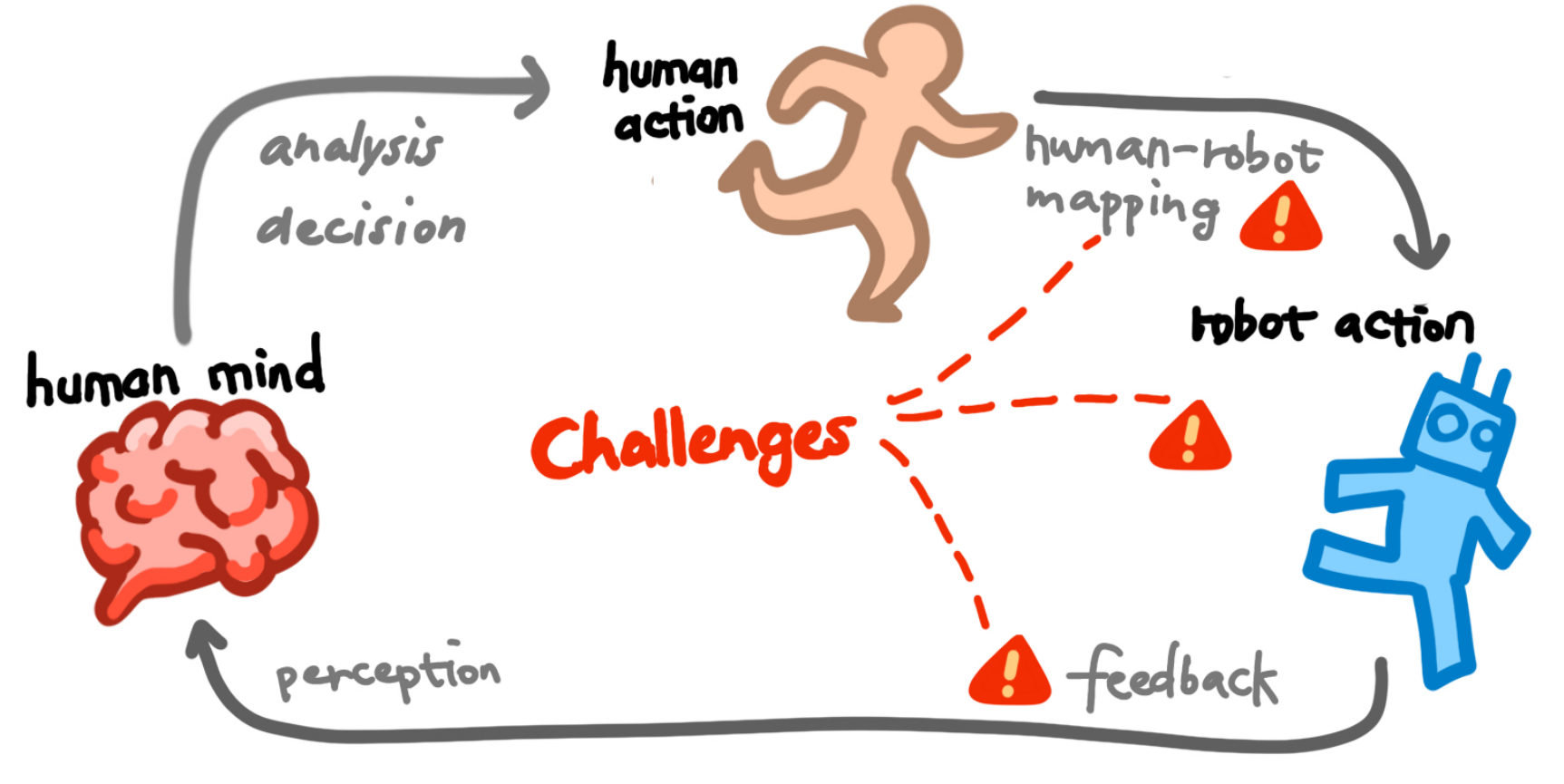}
    \caption{The human-robot interaction model in teleoperation. The process begins with the human brain analyzing prior knowledge and situational data to decide on a human action. Through the human-robot mapping, the robot responds with a corresponding action. Feedback from the robot's behavior is perceived by the human. The warning signs indicate HCI challenges at that step.}
    \Description{The human-robot interaction model in teleoperation. The process begins with the human brain analyzing prior knowledge and situational data to decide on a human action. Through the human-robot mapping, the robot responds with a corresponding action. Feedback from the robot's behavior is perceived by the human. The warning signs indicate HCI challenges at that step}
    \label{fig:cycle}
\end{figure}

In this iteration, we conceptualized the interaction model for a human operator performing robot teleoperation. We framed the design within a perception-action model (Fig.~\ref{fig:cycle}, which is a common concept shown in literature but takes various names and manifests with different compositions \cite{dominguez2022perception, pot2021perceived, geurs2004accessibility, cutsuridis2011perception}. In this model, the human-robot interaction process follows this sequence: the human brain processes prior knowledge and analysis to decide on an action, which leads to human action. This action, mediated by human-robot mapping, results in a robot action. The human then perceives feedback, which informs the brain for subsequent actions.

Prior knowledge is crucial in this loop. When humans have more prior knowledge of the task and system, they require less feedback (e.g., we move our hands freely without looking at them). Conversely, a lack of prior knowledge increases the importance of feedback to guide human’s understanding. As explained in Sec.~\ref{sec:challenges}, one major HCI challenge in robot teleoperation is understanding the mapping between the human body and the robot, given differences in their ranges of motion. This abstract mapping can lead to difficulty in controlling the robot, reducing the operator's ability to exploit the robot's full potential.

Given these considerations, our design aims to address three key issues: (1) maximizing the utility of an operator's prior knowledge, (2) providing clearer feedback, and (3) enriching the action space by introducing flexible mapping.

\begin{figure}
    \centering
    \includegraphics[width=0.5\linewidth]{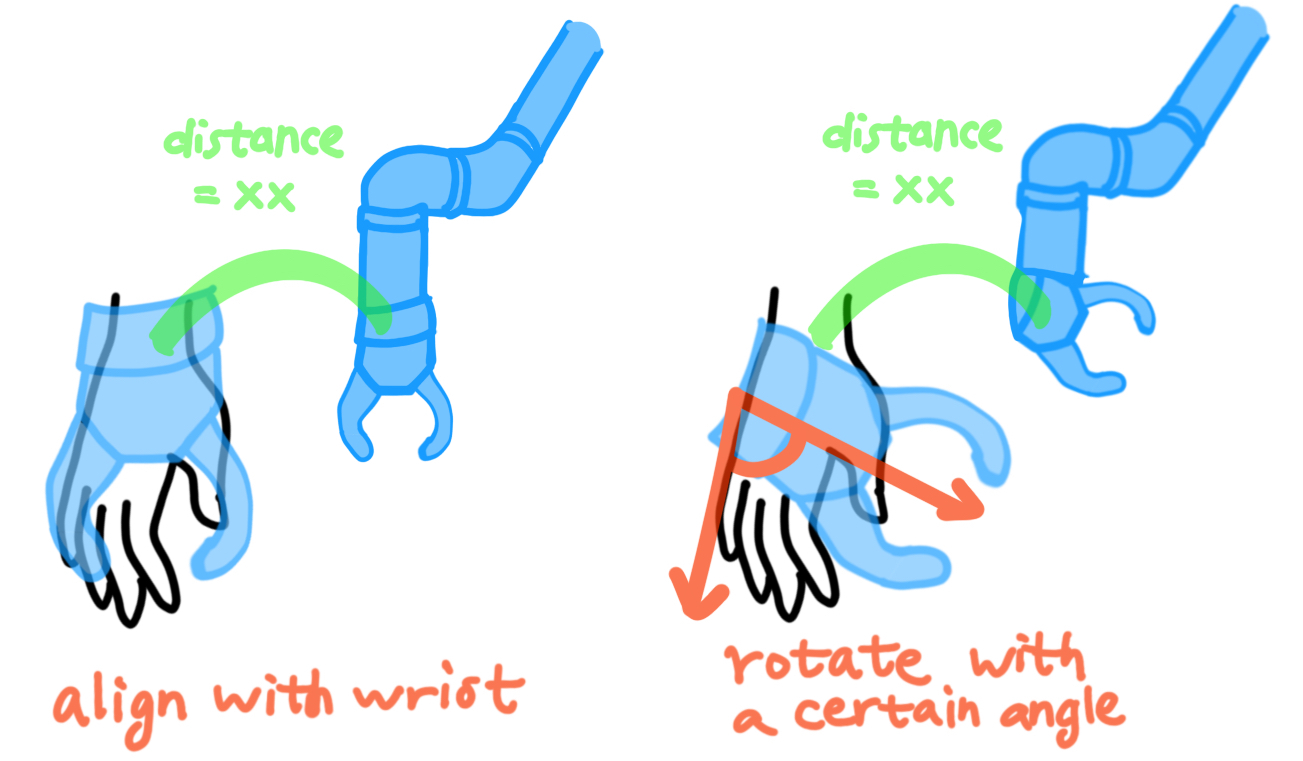}
    \caption{Embodied gripper with rotation offset in the first iteration of the design. The left side illustrates the robot gripper aligning with the user’s wrist, with a curved green line and distance number. The right side demonstrates the robot gripper rotating relative to the user’s wrist at a specified angle, preserving the same distance, allowing for varied and flexible interaction between the human and robot in teleoperation tasks.}
    \label{fig:iteration1}
    \Description{Embodied gripper with rotation offset in the first iteration of the design. The left side illustrates the robot gripper aligning with the user’s wrist, with a curved green line and distance number. The right side demonstrates the robot gripper rotating relative to the user’s wrist at a specified angle, preserving the same distance, allowing for varied and flexible interaction between the human and robot in teleoperation tasks.}
\end{figure}

\subsubsection{Maximizing the Utility of Prior Knowledge}
To make the mapping between humans and robots more intuitive, we embodied the robot gripper with the user’s hand. This embodiment allowed users to open and close the gripper simply by mimicking the action with their hand. Additionally, we matched the wrist position and hand orientation to the gripper base position and pose. This design leveraged the user's natural familiarity with their own hand movements, minimizing the learning curve and improving interaction efficiency.

\subsubsection{Providing Clearer Feedback}
To enhance feedback, we offer an option to overlay a semi-transparent virtual gripper on the user's hand in AR, ensuring it moved together with the user’s hand as though it were a skin. This visual cue allowed users to better understand how their hand movements translated to the robot’s gripper actions. To further clarify the correspondence between the hand and gripper, we resized the virtual gripper to better align with the hand's contour, enhancing the embodiment effect.

We also introduced a curved line (visual cues in AR) that connected the robot gripper and the hand, with numerical annotations showing the distance between the two. This helped users visualize the exact spatial relationship between their hand and the gripper during operation, providing clear and immediate feedback on the mapping.

\subsubsection{Enriching the Action Space}
To enable more flexible interaction, we introduced features that allowed users to manipulate the virtual gripper beyond simple direct mapping. Users could drag the virtual gripper away from its original pose to create position and orientation offsets. For instance, if the virtual gripper was rotated 90 degrees counterclockwise, moving the user’s hand forward would result in the gripper moving to the left. Additionally, users could adjust the scale of the mapping, so that moving their hand by one foot could move the robot by two feet if the scaling factor was set to 2. This enriched action space allowed users to customize the interaction, tailoring it to the task’s requirements and the robot’s capabilities.

\subsection{Iteration 2: Testing with Mid-Fidelity Prototypes in Simulation} 
% virtual robot arm. simulated time delay
In this iteration, we implemented the design in a virtual reality (VR) environment. We created a virtual robot with realistic motion characteristics, including time delay (around 1-3 sec depending on the complexity of movement and the motor speed setting) and kinematic accuracy, to closely simulate the behavior of the physical robot. This allowed us to test the effectiveness of the interaction techniques in a controlled, simulated setting. Users were asked to perform tasks in this environment to provide feedback on the design.

We invited two users (1F, 1M) for a 15-minute test session where they were asked to pick up virtual objects and move them around. Users quickly adapted to the embodied control and visualizations, however, users encountered several blockers that affected the overall experience:

\subsubsection{Time Delay and Feedback Confusion}
The simulated time delay, while realistic, introduced asynchronous feedback. This delay confused users, as the robot’s actions did not immediately align with their hand movements. Users found it difficult to synchronize their actions with the delayed feedback, leading to frustration and hesitation.

\subsubsection{Occlusion and Perception Challenges}
Another issue arose with visual occlusion, where users could not perceive the correct feedback due to parts of the virtual robot obstructing their view. For example, in the middle of an action, the arm portion of the robot may block the user's line of sight of the gripper. Another example is from some perspectives, the gripper may block the observation of the manipulated object. The occlusion made it difficult for them to fully understand how their actions affected the robot, further complicating the teleoperation experience.

\subsubsection{Rich Action v.s. User Confusion}
While the position offset feature was intuitive and easy for users to handle, the rotation offset caused significant confusion. Users struggled with the concept of moving the robot in directions that differed from their hand movements. They were comfortable with direct mappings, \ie moving in the same direction. Interestingly, they also found mirrored mapping  (180-degree flip) acceptable, since it leverages prior knowledge of a user seeing symmetric action in a mirror in daily settings. Other than the direct and mirrored mapping, users struggled with arbitrary rotation offsets. This misalignment between expected and actual motion disrupted the flow of interaction.

\subsubsection{Design Adjustments}
To address these blockers, we made several key adjustments to the interaction design:

\begin{itemize}[nosep,leftmargin=*]
    \item Freeze/Unfreeze Feature: We introduced a freeze/thaw feature to help users better observe and understand the robot's behavior. By freezing the robot, users could pause the teleoperation and move around to change their perspective. This allowed them to examine the robot’s position and assess its actions more clearly. Once they had a better understanding, they could thaw the robot and continue the operation seamlessly.
    \item Immediate Feedback via Virtual Robot: To mitigate the confusion caused by the time delay, we added a second virtual robot with no delay (in our design, the original virtual robot represented a physical robot with a realistic time delay). This secondary robot provided immediate feedback on how the robot would react to user actions in real time. By having this additional visual cue, users were able to anticipate the delayed robot’s movements, reducing time in trial and error and improving the overall experience.
    \item Simplifying Rich Action: Based on user feedback, we kept the scaling feature, which allowed users to adjust the mapping scale between hand movements and robot movements. However, we simplified the rotation offset feature, simplifying it into a mirror feature. Instead of allowing arbitrary rotations, users could now move the robot either in the same direction as their hand or in the opposite direction, creating a mirrored effect. This adjustment reduced confusion while maintaining flexibility in controlling the robot.
\end{itemize}

These design changes aimed to increase precision and reduce time in trial and error, and effort in adjusting while allowing for rich, flexible interactions with the robot.

\subsection{Iteration 3: Testing with High-Fidelity Prototypes on a Real Robot Arm} 

In the final iteration, we deployed the interaction design on a real robot (details in the implementation section later) and conducted a 30-minute pilot study with three users (2F, 1M, average age 28.3) from the research network. The users all had experience with XR, though only one had prior experience with robot programming. The goal of this study was to validate the interaction design in a real-world scenario.

Users were asked to relocate various objects on a tabletop, including a 6x6x6 cm cube, an 18 cm tall plush toy, and a 20 cm tall plastic bottle. The task did not have strict time constraints, and users were instructed to think aloud while performing the task. This allowed us to gather insights into how they interacted with the robot and whether any issues emerged during teleoperation.

While users did not experience confusion with the system features, several new issues specific to real robots were observed:
\textbf{Unexpected Jitter and Swing}
One of the primary issues was the occurrence of unexpected jitter and swing in the robot’s movements, which had not been seen in the virtual simulations. These behaviors were attributed to inconsistencies in the inverse kinematics (IK) solution, leading to erratic swings and potential collisions with the table. To address this, we leveraged the robot’s SDK to fine-tune the control parameters, ensuring smoother, more stable robot movements and preventing further collisions.
\textbf{Usability Improvements}
In addition to addressing the mechanical issues, users provided feedback on the visual feedback system. They found that the overlaid robot visualization overlay was too opaque, obstructing their view of the physical robot. Based on this feedback, we adjusted the transparency to a more optimal level (20\%), enhancing visibility without compromising the benefits of visual cues. Users also found the numerical values of distance more distracting than helpful, so we removed them. In addition, users suggested using a straight line rather than a curved line to connect the hand and the physical gripper, because it was easier to reflect the distance.

% With the identified issues—jitter, swing, and transparency—resolved, the interaction design was finalized. The pilot study demonstrated the feasibility of the system in a real-world setting, confirming that the interaction techniques were intuitive and usable for users with varying levels of robot experience.

% sensitivity of collision

% protection against table surface, protection of self collision

% how we smooth out the motion to decrease delay

\section{Arm Robot}

\subsection{Level of Embodiment in Teleoperation}
% Metaphors are the key concept of interaction design, especially for embodied interaction. Table \ref{tab:embodied_metaphors} depicts the metaphors we used in embodied control.

Arm Robot's interaction centers on hand embodiment, using the ``gripper as hand'' metaphor. However, we can use either a freehand, or handheld controller as the input modality. As we aim to investigate how much embodiment users appreciate in teleoperation, we design two versions of Arm Robot.

The first version is a freehand Arm Robot. Users' wrist movements are retargeted to the robot's gripper in 6-DOF, and hand openness is retargeted to gripper openness.
The other version is a controller-based Arm Robot. Users' hand pose, approximated by the controller's pose, is retargeted to the robot's gripper pose in 6-DOF, but the gripper openness is now bound to the trigger button on the controller. We hypothesized users may feel less sense of embodiment.

% The core interaction of Arm Robot is freehand embodied teleoperation. The metaphor is ``gripper is hand''. When one translates or rotates their wrist, the base of the robot gripper imitates the 6-DOF pose. When one's fingers close to "grasp", the robot gripper also grasps. 

\subsection{Visualization for Clear Feedback}

We display two robot visualizations rather than abstract text to augment user perception with visual feedback. 

\subsubsection{Temporal Discrepancy}
The first robot visualization is a full robot arm superposed right on the physical one. It shows the robot's motion without delay in AR. In other words, it is the zero-delay version of the physical robot. This robot visualization is made translucent to avoid any visual occlusion of the real robot. When the real robot arrives at the target pose indicated by the virtual robot, they will fully overlap. This intends to augment users' perception of different response times. 

This virtual robot also changes color from translucent white to bright orange when the intended motion is impossible due to robot kinematics. For example, if a user continuously moves in a direction and goes beyond the range of the robot's motion, the robot will turn orange as an anomaly signal. This feature aims to augment users' perception of different ranges of motion.

\subsubsection{Spatial Discrepancy}
The second robot visualization is a virtual gripper that overlaps with the embodying hand. The virtual gripper is translucent so users can still see their hand through the gripper. This visualization gives users an idea of the correspondence between their hand and the robot gripper.

\subsection{Interaction Techniques to Change Perspectives and Expand Action Space}

We adopted the absolute coordination system to track hand position. The human hand's range of motion can be extended by walking or bending. But this is not enough. 

We augmented users' range of motion through three interactions to change the spatial mapping between the hand and the gripper. They are the on/off switch, the scale of mapping, and the mirror reversal of mapping. We recommend turning off embodiment to free the hand for other operations like scaling or mirroring. 

\subsubsection{``Freeze/Unfreeze''}
We connected a straight line in AR between the wrist and the physical gripper. When the line is green, it means the embodiment is on. Users can use the other hand to pinch the line. After the pinch, the line will turn grey indicating that the embodiment is paused. We call it ``Freeze''. Once the gripper freezes, users can freely move to another position while the positional offset of spatial mapping keeps updated with the new relative position between the hand and the gripper. The new spatial mapping will take effect after the user pinches the line to unfreeze the gripper. Therefore, users can move around to observe from different perspectives without messing up the motion mapping.

% Users can extend their range of motion with this ``freeze - move - unfreeze'' interaction. Suppose one wants to move the gripper upwards to the robot extreme. In such a case, the user can first move the gripper to the highest possible position by fully extending their arm up via teleoperation. Then, they freeze the gripper there and move their hand from above their head back to chest level, which updates the spatial mapping between hand and gripper. When resuming teleoperation, the new spatial mapping takes effect and allows controlling the gripper above the initially reachable space by moving their hand from chest to overhead again. This technique extends the user's effective range without requiring exaggerated body movements like standing on a chair. 

\subsubsection{``Scale''}

Users can modify the scale of spatial mapping by interacting with a disk. They can resize the disk with two-hand manipulation. The disk radius indicates the ratio of robot motion to hand motion. For example, if the disk is resized as twice as large, one inch of hand movement will become two inches of gripper movement. The disk can be shrunk to half or enlarged to double. This feature is called ``Scale''.
This feature was designed to augment or shrink effective human motion range to adapt to the robot workspace.

\subsubsection{``Mirror''}

Mirror reversal of motion is designed as an alternative spatial mapping to fit into the motion range when the target object is between the operator and the robot. We place two solid arrows embodying the X and Y axes on the disk. Users can rotate an arrowhead 180 degrees to reverse the robot's motion on that axis. 
For example, when users approach the robot along this axis in ``Mirror'' mode, the robot will approach the users too, instead of shifting away in the same direction. The robot imitates the moves seemingly as the user's reflection in the mirror.

% We came up with these interactions by heuristics and common sense, knowing that they are not the exclusive designs to satisfy the design requirements.

% the chart of metaphor-virtual action-physical action: Eliciting Embodied Metaphors through Augmented-Reality Game Design

\begin{table}[htp]
    \centering
    \resizebox{\linewidth}{!}{
    \begin{tabular}{@{}p{4cm}p{7cm}p{6cm}@{}}
    \noalign{\hrule height 1pt}  
         \textbf{Metaphor} & \textbf{Virtual Action} & \textbf{Physical Action} \\
    \noalign{\hrule height 1pt}  
    \multirow{2}{*}{\textbf{Gripper is hand}} & Rotate and translate the gripper & Rotate and translate the embodying hand\\
     & Gripper grasps & Hand grasps \\
    \noalign{\hrule height 1pt}  
    Embodiment toggle is line & Pause/Resume embodiment & Pinch the line to switch between grey/green \\
    \midrule
    Scale is disk & Scale up/down the gripper translation & Drag the disk with two hands to resize it\\
    
    \midrule
    Coordinate axes are arrows & Reverse robot motion on the axis & Flip an arrow \\

    \noalign{\hrule height 1pt}  
         
    \end{tabular}}
    \caption{Metaphors, virtual actions, and physical actions of embodied interactions in Arm Robot.}
    \Description{
    This table presents metaphors for embodied interactions in Arm Robot, with corresponding virtual and physical actions. The “Gripper is hand” metaphor equates rotating and translating the gripper with similar hand movements and associates the gripper's grasp with the hand's grasp. “Embodiment toggle is line” metaphor links pausing and resuming the embodiment to pinching a line that toggles color from grey to green. The “Scale is disk” metaphor relates scaling the gripper's movement to dragging a two-handed disk to resize. Lastly, the “Coordinate axes are arrows” metaphor connects reversing the robot's axis motion to flipping an arrow. 
    }
    \label{tab:embodied_metaphors}    
\end{table}

\section{Implementation and Setup}
\label{system}
\begin{figure}
    \centering
    \includegraphics[width=\linewidth]{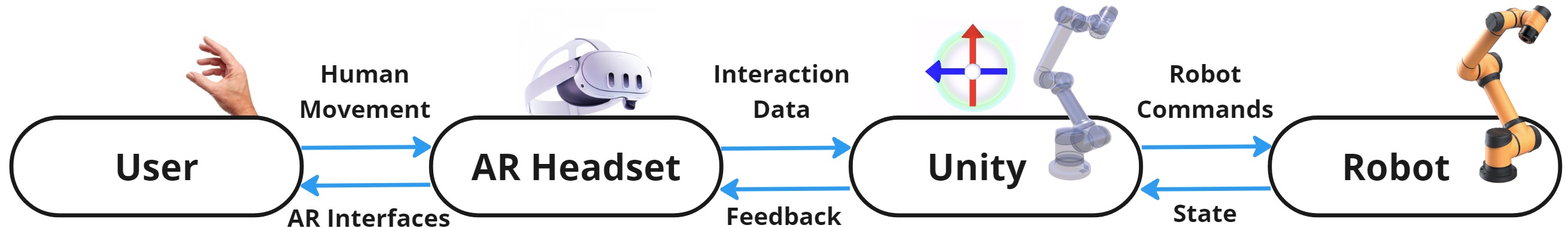}
    \caption{The high-level system architecture of Arm Robot shows the data communication between the user, AR headset, Unity, and the physical robot.}
    \Description{This flowchart illustrates the system architecture of an AR-enhanced robot control setup. It starts with a 'User' hand gesture, flowing into an 'AR Headset', through 'Unity' software, and culminating in a 'Robot' arm. Data communication includes human movement, AR interfaces, interaction data, feedback, and robot commands, with Unity as the central processing hub.}
    \label{fig:architecture}
\end{figure} % unfold the chain into a line. put both physical and virtual robots in the robot block
% do not need to add a streaming TV, and PC. draw the user as a circle (head) and shoulder

\subsection{Implementation}
The system of Arm Robot consists of an AR headset for tracking and rendering, a Unity project for embodied interfaces and visualization, and a physical robot arm with a gripper. Specifically, we used Meta Quest 3 as the AR headset, which supports passthrough mode and hand tracking with built-in cameras. The AR headset was connected to a PC (with an AMD Ryzen 7 3700 CPU and an RTX 3070 8G GPU) using a 20-feet USB 3.0 Type-C cable to offload the computation. In the Unity Engine (2023.1.9f1 version) installed on that PC, we prototyped the robot visualization.

Our robot, AUBO i5 series, is a 6-DOF robot arm with maximum linear speed at \SI{2}{m/s} and maximum reach at 886.5 mm. The gripper is DH 145, with a maximum openness of 145 mm. The visualization of the robot was created using the URDFs of the AUBO robot arm and the DH gripper by joining the arm and the gripper base. URDF is a file type that includes the physical description of a robot, widely used in realistic simulation. The robot visualization has the same structure and range of motion as the physical one. We can control the digital and physical 6-DOF robots in the same way by sending a valid command of six joint angles, which specifies the rotation of each joint motor to uniquely determine the robot pose. By ``valid'' it means the target robot pose is physically feasible for the robot. The only difference between digital and physical robots is the response time. The robot visualization can respond to user action with zero delay. The physical robot always has delay because the physical components of a robot arm, like gears, joints, and actuators, have inertia and thus take time to start moving or change direction. In addition, impedance control of the physical robot involves force and torque sensor reading, which introduces latency in sensing and communication. 

The AR headset sends the position and orientation of the wrist to Unity, and Unity computes the inverse kinematics solutions to generate a command of joint angles. The IK solver we used in the prototype is BioIK \cite{starke2016efficient, starke2017evolutionary}. Given a 6 DOF target pose, BioIK searches for valid joint angles that reach the target. We configured it to iterate 3 generations per frame (35 FPS) and evolve 120 potential solutions per generation. Only one successful solution was picked. We also set the smoothing parameter to 0.5, blending the previous joint positions and the current for a smoother transition between postures.
In addition, the target configuration was immediately assigned regardless of realistic acceleration and velocity, because the robot visualization should serve as a zero-delay preview of the physical robot. The same values of joint angles will be sent to the physical robot instantly. As a result, users can always see the physical robot trying to reach the digital robot's pose with the same IK solution.

\subsection{Setup}
\begin{figure}[htp]
    \centering
    \includegraphics[width=0.4\linewidth]{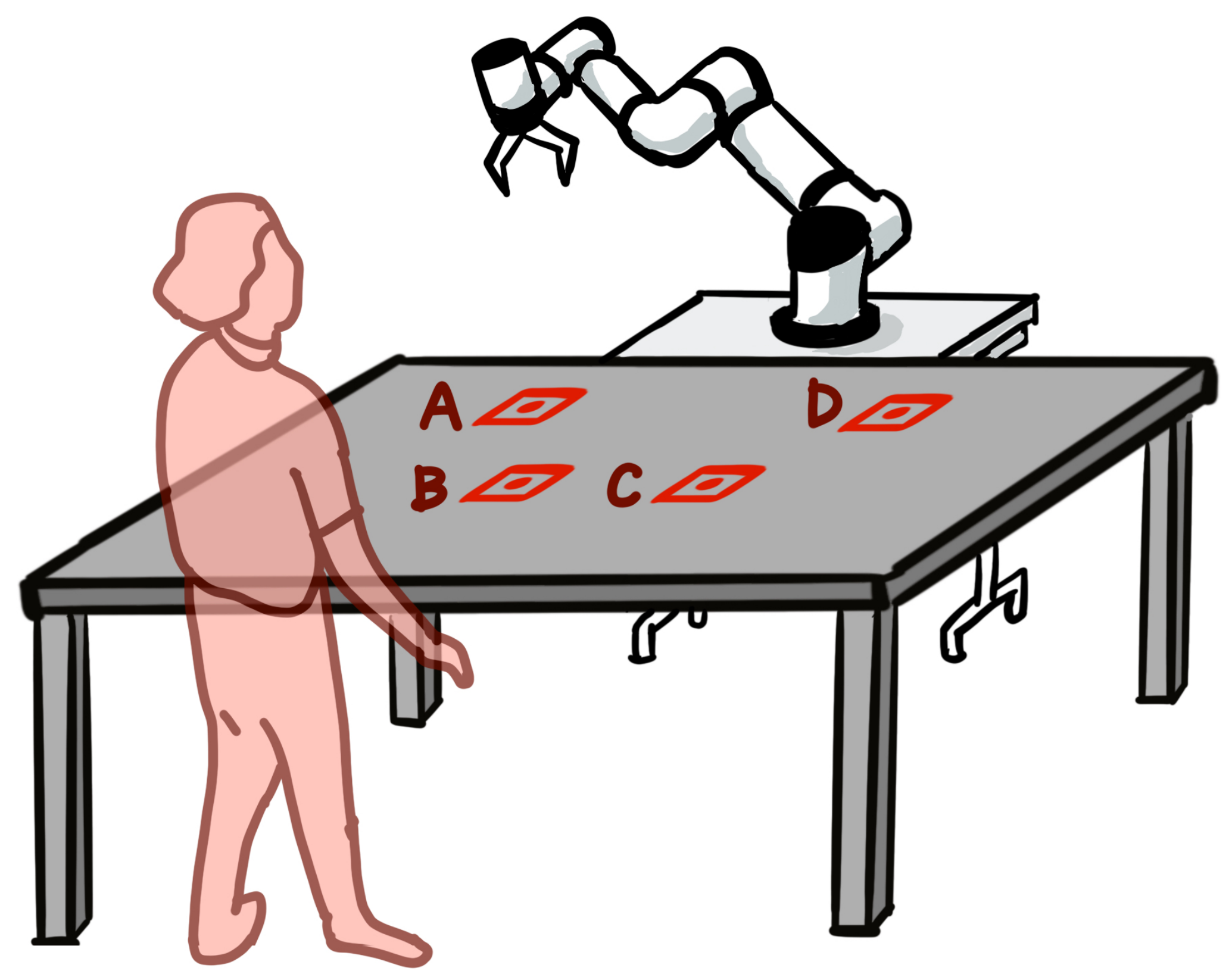}
    \caption{In our study, the robot is mounted beside the table, and the marker locations (A-D) are the start and end positions in evaluation tasks. A user can stand anywhere around the table to complete tasks.}
    \Description{
    In the center is a large rectangular table with the word ``Table'' below it. Above the table, a schematic of a robot is positioned with the label ``Robot'' just above it, indicating its location. Four red squares labeled A, B, C, and D are placed around the table, presumably representing marker locations for evaluation tasks. Below the table, an oval labeled ``User'' represents the user activity area, suggesting where the participant would be positioned during the interaction.
    }
    \label{fig:floorplan}
\end{figure}

We also need to calibrate the orientation and align the coordination system origins to superpose the virtual robot at the same location as the physical one. As shown in Fig.~\ref{fig:floorplan}, the physical robot was mounted on a desktop (600 mm wide, 1150 mm long, 700 mm high). The desk was placed adjacent to a tabletop (1050 mm wide, 2100 mm long, 750 mm high). We placed a marker on the floor during initialization. The distance shift between the physical robot and the marker in the real world is used to initialize the position of the virtual robot to the headset location on the XoY plane in Unity. The headset in Unity faces the direction of the Y+ axis. The Z-axis elevation is set to desktop height. In this way, as long as users stand on the marker, and face the Y+ axis when entering the AR app, the coordination system in AR will align with the one in reality.

\section{User Study}
With feedback from the pilot study, we improved the Arm Robot and conducted a user study to further investigate RQ2 and RQ3. 

\label{userstudy}

\begin{figure}[htp]
    \centering
    \begin{subfigure}[b]{0.6\textwidth}
       \includegraphics[width=\linewidth]{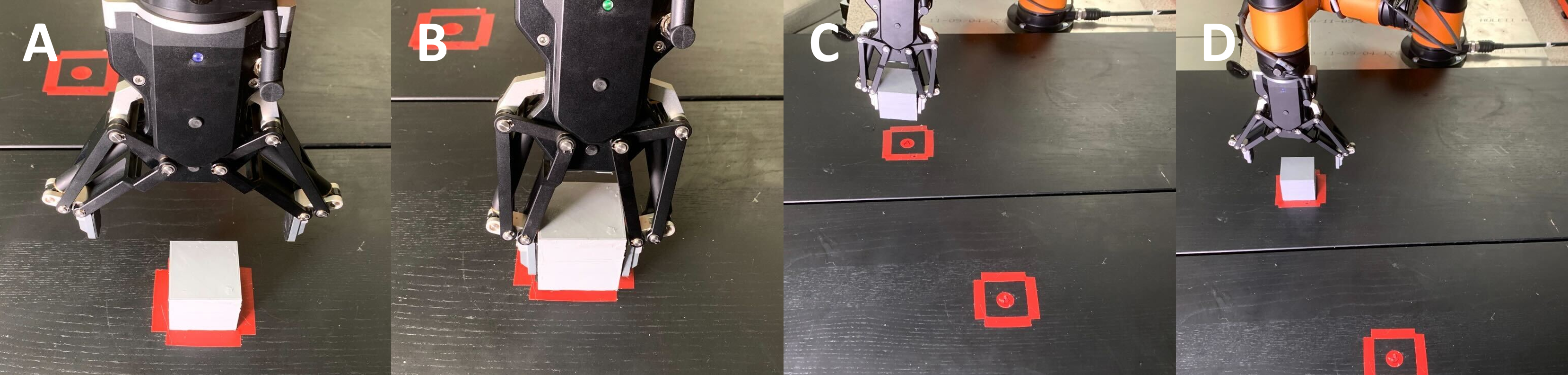}
       \caption{}
       \label{fig:cubetranslation} 
    \end{subfigure}
    \hfill
    \begin{subfigure}[b]{0.6\textwidth}
       \includegraphics[width=\linewidth]{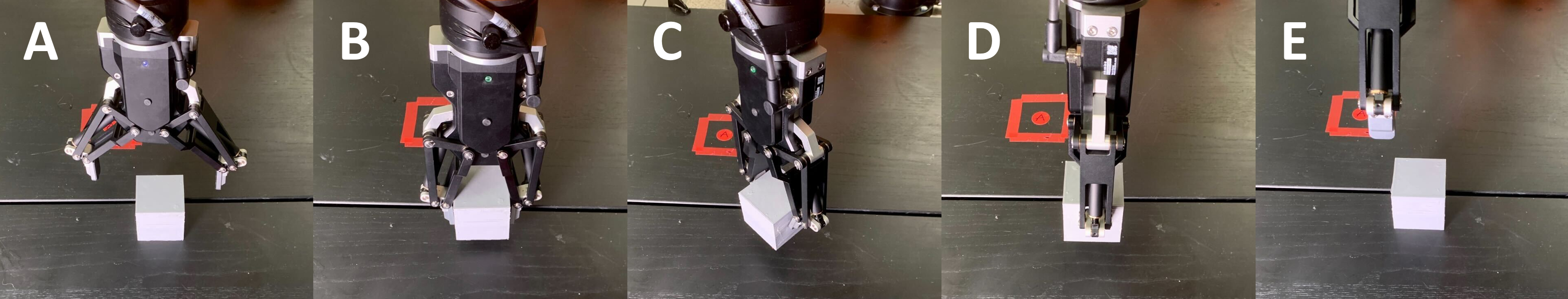}
       \caption{}
       \label{fig:cuberotation}
    \end{subfigure}
    \caption{Performance evaluation with a cube in (a) translation, and (b) rotation.}
    \Description{A sequence of a robotic gripper conducting performance evaluation with a cube. In set (a), the gripper translates the cube across four stages, A to D, with precise linear movement. In set (b), the gripper rotates the cube through five stages, A to E, demonstrating a controlled rotational manipulation.
}
    \label{fig:evaluationtasks}
\end{figure}

\begin{figure}[htp]
    \centering
    \begin{subfigure}[b]{0.85\textwidth}
       \includegraphics[width=1\linewidth]{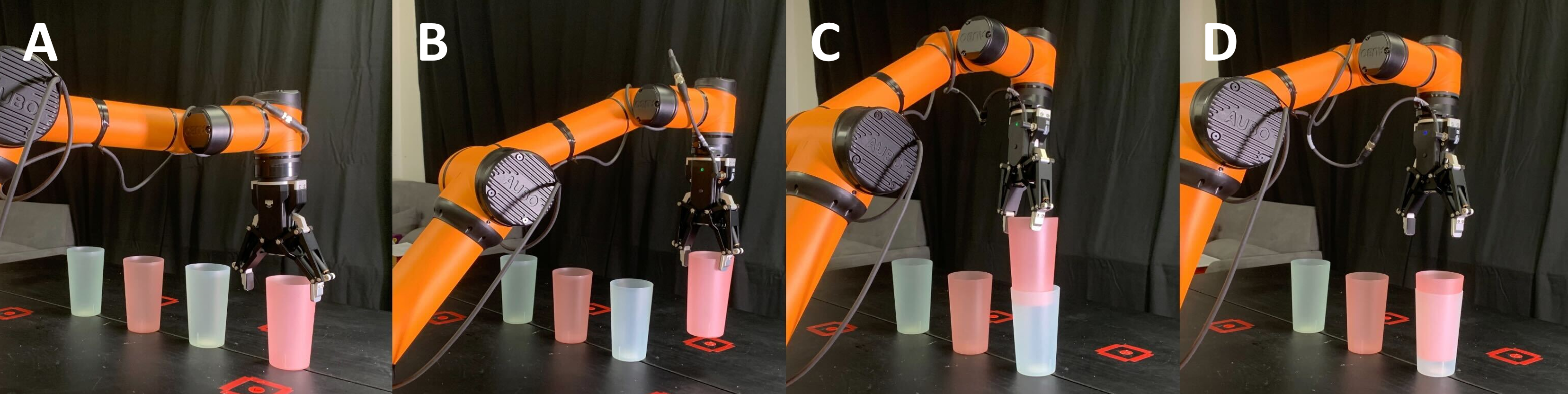}
       \caption{}
       \label{fig:cupstackingtasks} 
    \end{subfigure}
    
    \begin{subfigure}[b]{0.85\textwidth}
       \includegraphics[width=1\linewidth]{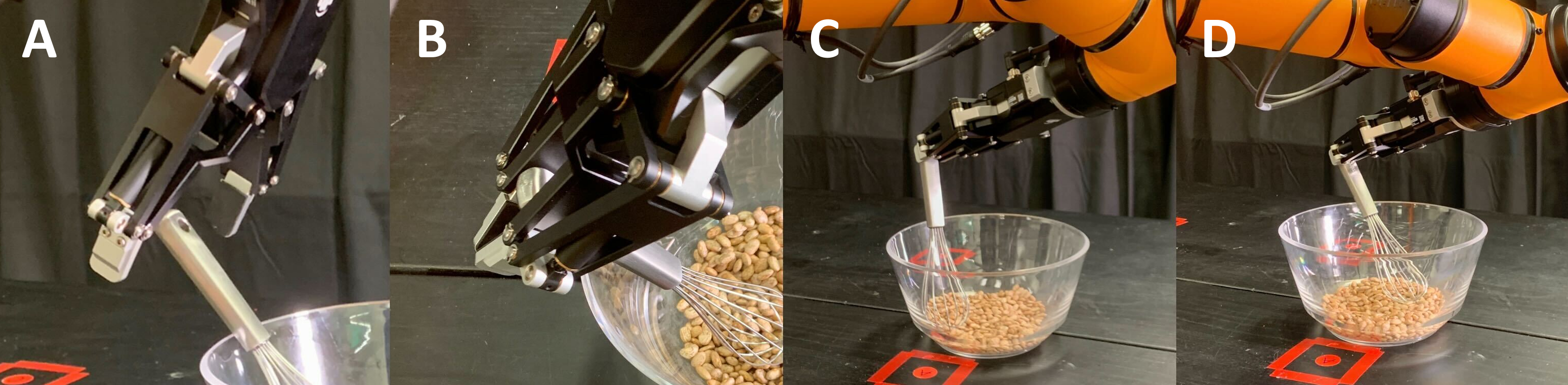}
       \caption{}
       \label{fig:whiskingtask}
    \end{subfigure}
    
    \begin{subfigure}[b]{0.85\textwidth}
       \includegraphics[width=1\linewidth]{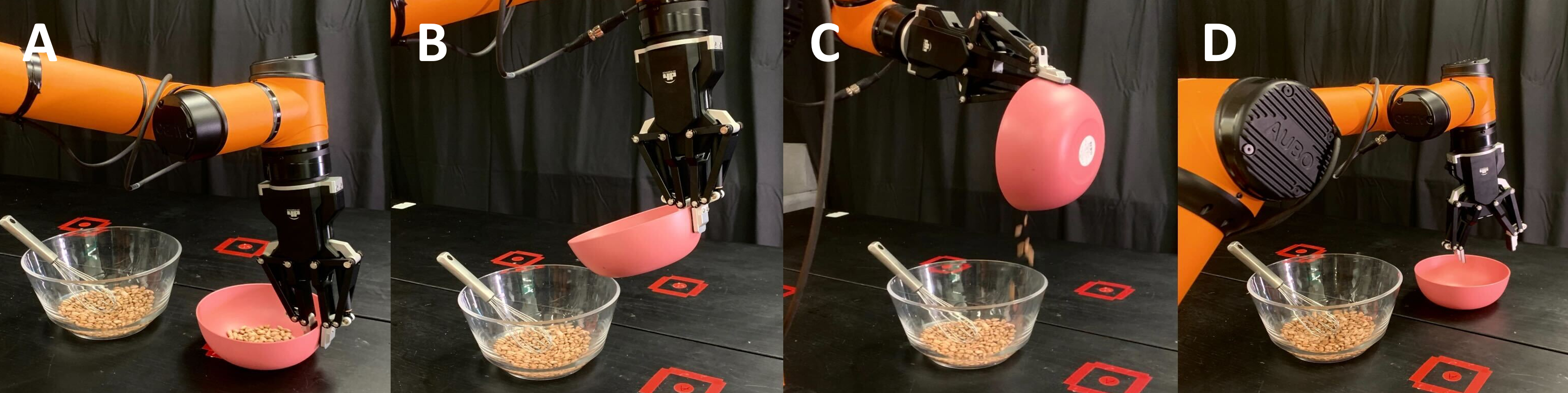}
       \caption{}
       \label{fig:pouringtask} 
    \end{subfigure}
    
    \begin{subfigure}[b]{0.85\textwidth}
       \includegraphics[width=1\linewidth]{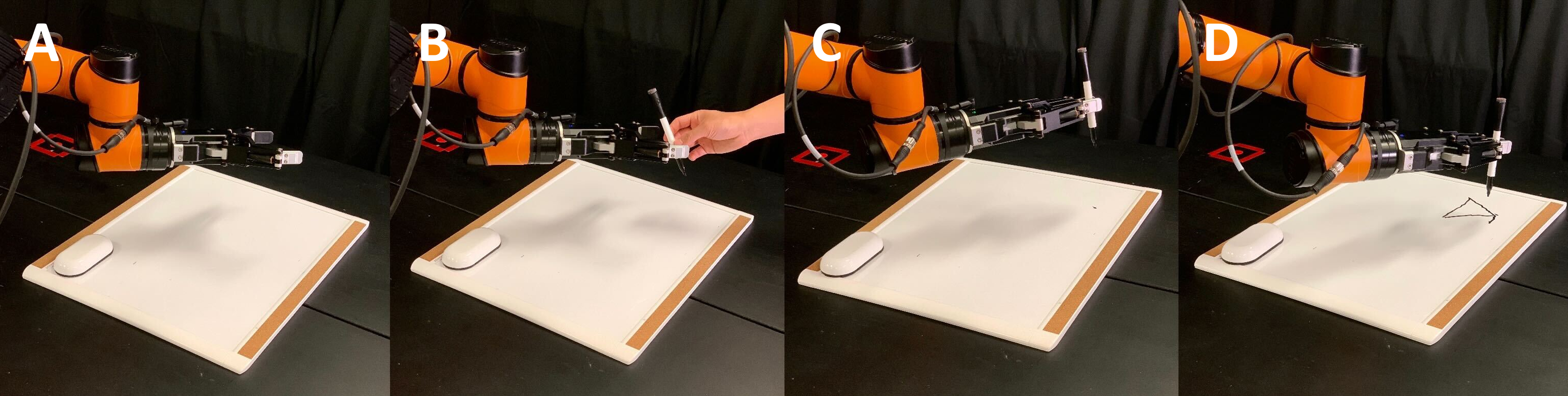}
       \caption{}
       \label{fig:writingtask}
    \end{subfigure}
    
    \begin{subfigure}[b]{0.85\textwidth}
       \includegraphics[width=1\linewidth]{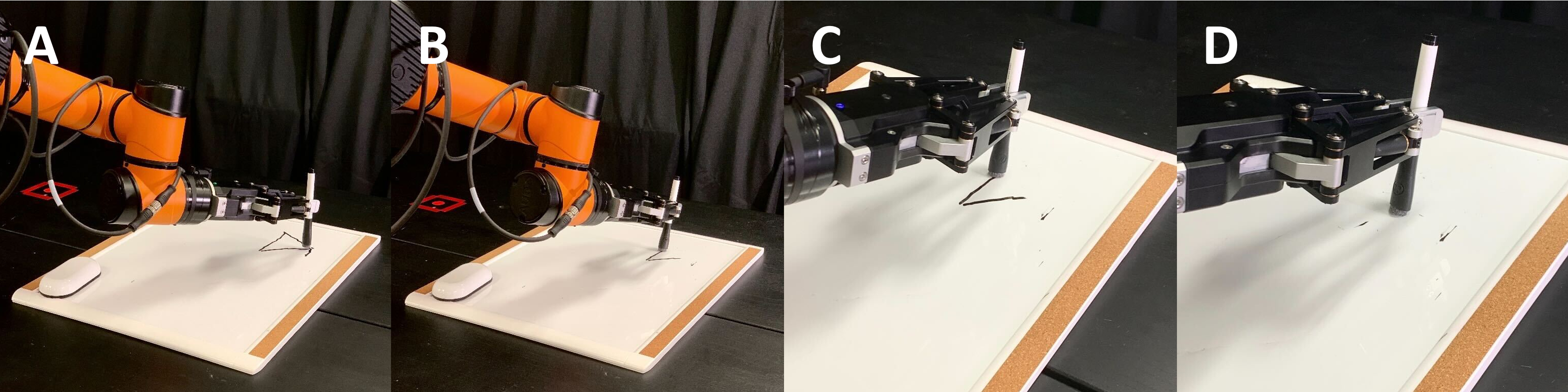}
       \caption{}
       \label{fig:erasingtask}
    \end{subfigure}
    \Description{Photo series displaying a robotic arm performing various tasks with daily objects. Sequence. (a) shows cup stacking with two colored cups. Sequence (b) captures the whisking of ingredients in a bowl. In sequence (c), the robot is pouring beans. Sequence (d) and (e) depict the robot drawing with a pen and erasing on a whiteboard.}
    \caption{Exploration with daily objects. Tasks include (a) cup stacking (b) whisking (c) pouring (d-e) drawing/erasing }
    \label{fig:explorationtasks}
\end{figure}
% use a sequence instead of one photo, to show pick up - move- release

\subsection{Participants}
We recruited 18 participants (9F, 9M), aged 21-29 (M = 23.67, SD = 2.47) from undergraduate and graduate students for the in-person study. network. Their backgrounds span across Biomedical Science, Computer Science, Psychology, Law and \etc. While most had no robot programming experience, four participants had some, varying from occasional to daily programming. AR/VR technology usage was more common, with 10 participants reporting experience. The study was approved by the Institutional Review Board. Details are in Table \ref{tab:demographics} of Appendix.

\subsection{Two Variants of Arm Robot}

To evaluate the degree of human body engagement as a factor to usability, we created two versions of Arm Robot to compare, by replacing freehand interaction with controller-based interaction. With a controller, the robot gripper was controlled by a trigger button. Pinching the line was replaced by pressing another button. When the embodiment was turned off, the trigger button was used to simulate hand grabbing the disk and arrows for scaling and mirroring. In a word, the less embodied Arm Robot still had adjustable embodiment and visualization, and let users embody their motion, but with less body engagement. The interactions for spatial correlation adjustment had also become more abstract.

While the study compared the freehand Arm Robot with the controller-based Arm Robot, \textbf{our purpose is not a performance competition, but to answer RQ3 -- How the extent of body embodiment affects the usability of embodied robot arm control. }We aim to extract findings about how much body engagement users want and dive into their reasoning.

In addition, the comparison between the freehand Arm Robot and the controller-based Arm Robot is not the only goal of our study. The study also dived into RQ3 -- the adjustable spatial correlation in robot arm manipulation by analyzing when, why, and how users would modify the spatial correlation.

\subsection{Tasks}
Our Tasks are divided into Evaluation and Exploration. Users could freely adjust the spatial correlation of embodiment when doing the following tasks.

\subsubsection{Evaluation with a Cube}
This task aimed to quantify the performance of the Arm Robot with various degrees of body engagement.

\begin{itemize}[nosep,leftmargin=*]
    \item \textit{Translation:} Participants were instructed to move a cube from one square marker to another, as illustrated in Figure \ref{fig:floorplan}, which shows the layout of four squares on a tabletop. A successful trial required picking up the cube from its original marker, translating it to the target marker without rotation, and then releasing the gripper to place the cube on the target marker (Figure \ref{fig:cubetranslation}). Participants were randomly assigned a pair of markers as the start and end positions. They were required to perform the task until three successful trials were achieved. Participants counted down from three when the gripper was about one inch above the cube, then we started timing. We ended the timing upon releasing the cube. Participants aimed to complete fast.  
    
    \item \textit{Local Rotation:} This involved lifting the cube slightly above the tabletop, rotating it 90 degrees on the XoY plane without translating it, and then releasing it at the same location (Figure \ref{fig:cuberotation}). The cube was positioned between the start-end marker pair used in the translation task, and aligned parallel to the table edge.
\end{itemize}

For both the translation and rotation tasks, users needed to finish them with a freehand Arm Robot and a controller-based Arm Robot. The order of the Arm Robot and controller-based Arm Robot was randomized for each user.

\subsubsection{Exploration with Daily Objects}
In this session, participants further explored interaction with daily objects through a series of comprehensive tasks using the Arm Robot. Completion of all tasks was encouraged but not mandatory, and tasks were not timed. Users could freely adjust spatial correlation. 

\begin{itemize}[nosep,leftmargin=*]
    \item \textit{Cup Stacking:} Participants were presented with four plastic cups placed in a row with intervals between them. Successfully picking up one cup and inserting it into another without falling was considered a success.
    \item \textit{Whisking:} A mixing bowl and a whisk were provided. Participants embodied the robot to blend food ingredients in the bowl with the whisk. Success was defined as moving the whisk in a circle without spilling content.
    \item \textit{Pouring:} Participants were asked to transfer food from its container to a mixing bowl by tilting the container and pouring the content into the bowl. No spilling was a success.
    \item \textit{Drawing/Erasing:} A whiteboard and marker were provided, where the marker had an eraser on the opposite end. Participants were asked to draw and then erase their creations, with success defined as completing more than one continuous stroke for either drawing or erasing.
\end{itemize}

\subsection{Measures}

For evaluation with a cube, we recorded task completion time, and noted down the number of attempts before each successful trial. Additionally, we conducted semi-structured interviews to obtain Likert-scale scores for usability and dive deeper into their reasons. Specifically, we used six usability metrics on a 7-point Likert scale, with freehand Arm Robot or the controller-based Arm Robot. 1 stands for ``strongly disagree'', while 7 stands for ``strongly agree''. The six statements are as follows, 
\begin{itemize}[nosep,leftmargin=*]
    \item \textbf{Ease of learning:} The interactions are easy to learn and grasp.
    \item \textbf{Ease of recollection:} The interactions are easy to remember.
    \item \textbf{Effectiveness:} The interactions' functions effectively support me in completing all tasks.
    \item \textbf{Comfort:} The interactions are comfortable.
    \item \textbf{Efficiency:} The interactions allow me to complete all tasks fast.
    \item \textbf{Satisfaction:} Overall, I am satisfied with the interactions.
\end{itemize}

For exploration with daily objects, we observed how they embodied and guided the robot in finishing the tasks, noted their preference for body engagement, and logged their adjustment of spatial correlation in embodiment. We followed up with semi-structured interviews to better understand them.

\subsection{Procedures}
The study started with an introduction to the Arm Robot's goals and study protocol. Participants then filled out a demographic survey.
Based on their prior experience in robot programming and XR technologies, participants received brief or detailed tutorials on using the AR headset and the robot arm. After tutorials, we demonstrated each interaction in Arm Robot. 

After participants put on the AR headset, they had a 10-minute playground session to familiarize themselves with the freehand and controller-based versions of Arm Robot. Throughout the study, users were free to move around and stand anywhere in the space, \eg in the front, or by the side of the table.
When they were ready, we started the evaluation with the cube. The order of the freehand Arm Robot and controller-based Arm Robot was randomized. The start and end positions of cube translation were also randomly picked from the following: A to B, B to C, C to D, D to C, C to B, and B to A. This session took around 20 minutes.

After the evaluation tasks, users started exploration with daily objects. Users could freely adjust the spatial correlation of embodiment. They could also choose their preferred body engagement level -- choosing between controller-based and freehand Arm Robot. This session lasted around 20 minutes.
Once participants finished all tasks, study coordinators started the semi-structured interviews on usability with the Likert scale and open-ended questions (Appendix). We thanked participants with a compensation of 20 USD/hr.

\section{Results and Findings}
\label{findings}
\subsection{Performance}
\begin{figure}
    \centering
    \includegraphics[width=0.5\linewidth]{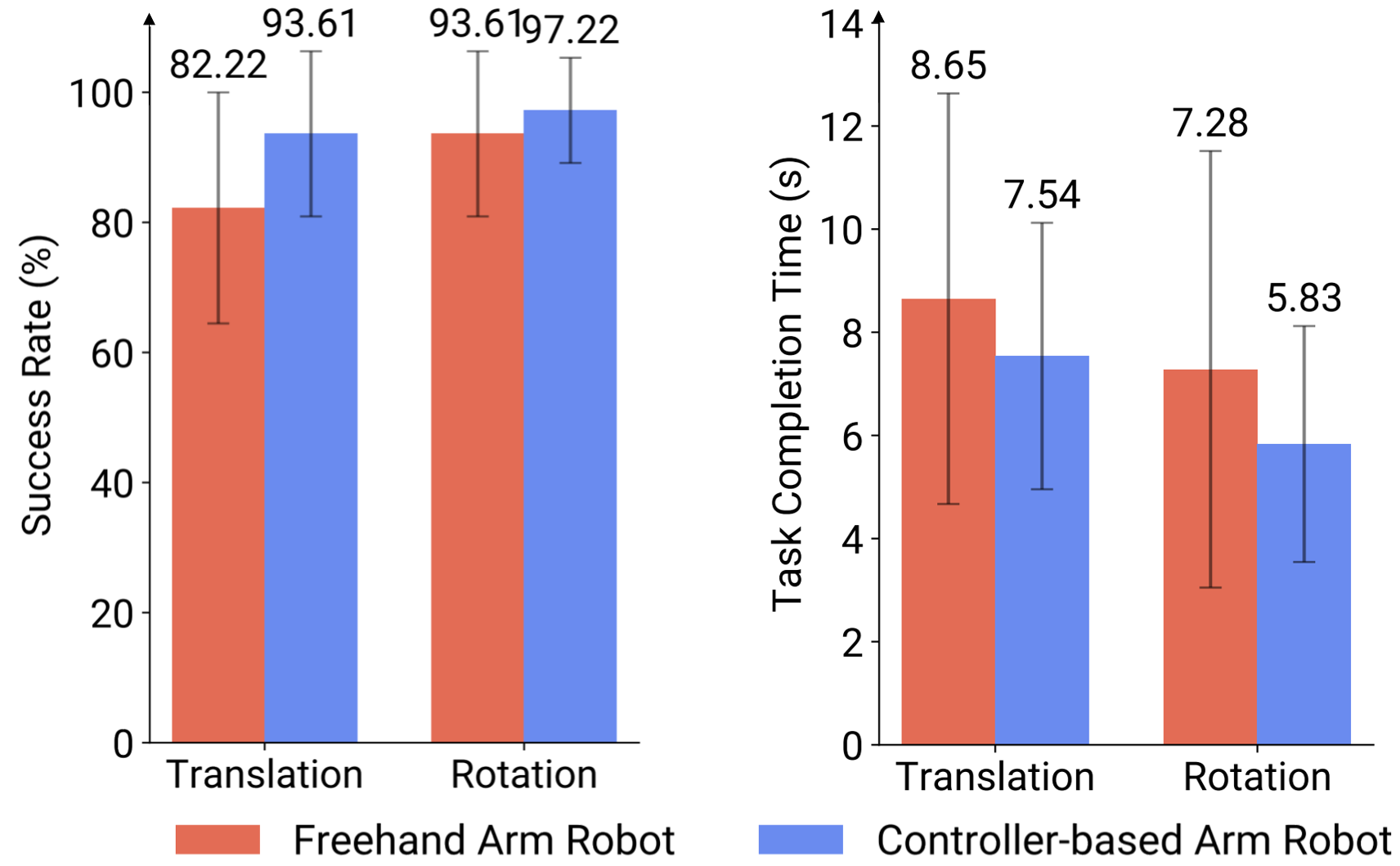}
    \caption{Completion time (s) and success rate (\%) of evaluation tasks using freehand Arm Robot and controller-based Arm Robot.}
    \Description{Bar graphs comparing the performance of freehand Arm Robot and controller-based Arm Robot. The left graph shows success rates, with the freehand method scoring slightly lower than the controller-based for translation and rotation tasks. The right graph shows task completion times, with the freehand method taking longer to complete tasks. Both robots exhibit high efficiency and accuracy, with the controller-based robot having a slight edge in both categories.}
    \label{fig:eval_result}
\end{figure}

\begin{figure}
    \centering
    \includegraphics[width=0.7\linewidth]{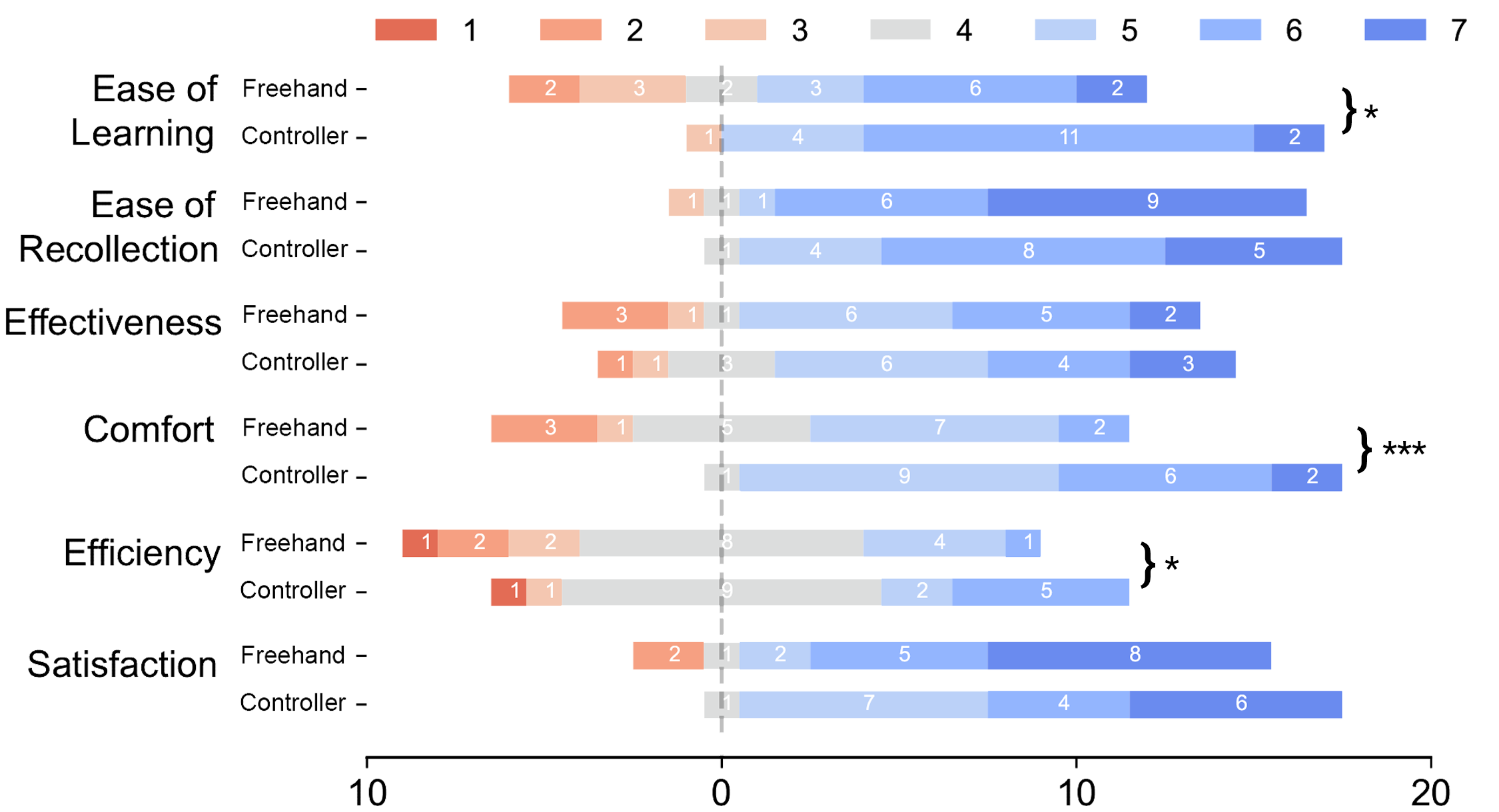}    
    \caption{A diverging stacked bar chart shows user responses on a 7-point Likert scale for usability metrics comparing freehand and controller-based Arm Robots. Categories evaluated include ease of learning, ease of recollection, effectiveness, comfort, efficiency, and satisfaction. Statistical significance is marked with asterisks in some categories, suggesting differences in user experience between methods.}
    \label{fig:likert_scale}
    \Description{A diverging stacked bar chart shows user responses on a 7-point Likert scale for usability metrics comparing freehand and controller-based Arm Robots. Categories evaluated include ease of learning, ease of recollection, effectiveness, comfort, efficiency, and satisfaction. Statistical significance is marked with asterisks in some categories, suggesting differences in user experience between methods.}
\end{figure}

% how the extent of body embodiment affects usability (quantitative)
Figure \ref{fig:eval_result} shows the completion time of translation and rotation tasks in evaluation. We took averages across individuals to address the order effect of methods. The average duration of translation and rotation is 8.65 s and 7.28 s for the freehand Arm Robot, and 7.54 s and 5.83 s for the controller-based Arm Robot. With handheld controllers, participants performed tasks significantly faster than freehand Arm Robot in rotation tasks (p-value = 0.014).
Figure \ref{fig:likert_scale} presents the score distribution across usability metrics on a 7-point Likert scale. We performed a significance analysis on the Likert scale between the two methods using a paired one-tailed T-test. Significant differences were found in ease of learning (p-value = 0.012), comfort (p-value = 0.00027), and efficiency (p-value = 0.018).

% ADD A BAR CHART FOR COMPLETION TIME

% \subsubsection{Common Pattern in User Behaviors}
% In the first few trials, we noticed that all users paid intense attention to the robot and moved very slowly to see if the motion was expected. U1 expressed why she was so focused because she was building a mental model of how this robot should work. She mentioned the discrepancies between humans and robots as barriers and how she tried to overcome them, ``If I were teaching (the robot), I need to learn its ability and limitations, like its scope... I will try to think from the perspective of the robot arm itself. So I'm helping it plan its optimal route to get that target.'' 

% design implications
% Results show Arm Robot helped users address the discrepancies in teleoperation.
% The usefulness of spatial mapping parameters in embodied teleoperation ranked as follows, from most useful to least useful: on/off, scale, and mirror reversal.

\subsection{RQ2 Findings on Visualization: When and why would participants look at the visualization of the robot arm?} % digital twins = "visualization" in the title
% use this framework to understand how xx improved usability.
% when -- when did users want to use this interaction/visualization. Did designers expect all the motivation?
% How -- How do they use this interaction/visualization to address the challenge? Was it expected in the design phase?
% Pay attention to the surprising parts.

17/18 participants regarded the digital robot arm as useful. As expected, they leveraged it as a real-time preview of the robot's target pose. They could determine whether the robot had reached the destination by telling if the physical robot overlapped with the digital one. Many gave high praise, \eg P16 stated ``Without that, I cannot tell how much it is moving or where it is moving. The virtual robot significantly improved the accuracy of finding out what was going on.'' The only user who disliked it was P6. This was because he squatted and walked around frequently to change perspectives, which challenged the robustness of head tracking. After some time, the tracking error accumulated and shifted the robot visualization's location, which made the predictive path unreliable.

We noticed that participants leveraged it in various ways. We summarize the patterns as below: % How often

(1) Looked at the robot visualization \textit{conditionally}. For example, P7 commented ``I don't look at virtual robot/gripper, unless when there is misalignment [in robot appearance].'' This is because P7 moved slowly where latency was trivial. She only looked at the physical one when it went out of range and turned orange for warnings.

(2) Switched attention between the robot visualization and the physical robot \textit{strategically}. For instance, P12 shared his tip to balance precision and speed, ``I initially look at the virtual robot, once I am near an object, I look at the real arm for precision.'' P13 also had a similar division, ``for translation, I mainly look at the virtual robot and wait for the physical robot to arrive. For rotation, usually the lag is not that obvious so I can directly look at the gripper.'' P15 also mentioned ``As soon as I need to position or grip it, I would look at the real one.'' We were surprised to see how fast users developed their strategies with available features.

(3) Looked at the robot visualization \textit{all the time}, with an occasional glance of the physical robot. For example, P18 deeply trusted our robot visualization and even ignored the physical robot when performing tasks. Fortunately, our robot visualization was accurate enough for predictive display.

Another interesting observation -- People tend to underestimate the inconvenience caused by latency. For example, P11 did not think robot visualization would be useful although she paid attention to our study introduction. However, after realizing the importance of robot visualization through hands-on experience, she completely changed her attitude towards it. She said, ``I didn't understand why it was there at the beginning, but after I started translating, I only looked at the virtual arm instead of the real one. After understanding how to use it, it helped a lot.'' 
Besides, participants perceived the virtual gripper overlaying on hand as being helpful. They easily understood the correspondence through embodied visualization.

\subsection{RQ2 Findings on Adjustable Spatial Mapping: When and why would participants change spatial mapping?}

\subsubsection{Freeze/Unfreeze -- Pause/Resume Embodiment.} 

All users reported that the ability to ``Freeze/Unfreeze'' was necessary and were using it frequently in sequences characterized by freeze, change location, and then resume. 
While this feature was designed to endlessly extend the range of motion or let users take a break, we noticed users would use this feature for other purposes. 

One application that we did not expect to be popular turned out to be heavily used by participants -- change viewing angles. This was observed among all participants. People did that for various reasons. P6 even squatted down to observe the distance between the gripper and the cube to ensure a successful grasp. Because Any obstruction to the gripper may influence task performance, most people tried various viewing angles before they found the sweet spot. 

While many other users preferred standing in the front for best visibility of the task, one user preferred standing behind or beside the robot for a stronger sense of embodiment. P18 froze, moved, and unfroze until he found the sweet spot that delivered the strongest sense of embodiment. He said, ``Standing behind/to the side would be easier since I can see it from the robot's perspective.''

Another important use case was to free the embodying hand for other tasks in the real world. For example, participants froze the embodiment so that they had spare hands to scale up the disk.
In addition, some participants suggested adding features of only freezing translation, or rotation, or motion on a certain axis. P8 pitched that, with these future features, ``it would be easier to do local rotation''.

\subsubsection{Scale -- Change the Scale of Embodied Motion.}
15/18 participants found ``Scale'' nice to have. We designed this feature to extend or shrink the effective motion range of the hand, and found participants used it for different purposes. The unexpected new usage of ``Scale'' mode included,

(1) Speed control. As P11 interpreted, ``Scaling was necessary because controlling the speed of things [as they were being moved] was very important.''

(2) Visibility and comfort. For example, P8 wanted to secure the standing location at the robot's front for an unobstructed view of the gripper. Therefore, he scaled up the motion to comfortably control the robot without exaggerated hand movement, standing at the same location.

(3) Stability and precision. We found that participants scaled down to have higher stability and precision. P16 said, ``Sometimes I need precision work so scaling down is very necessary.'' In this line of thought, P14 felt high scaling did not help him much mainly due to safety concerns (due to possible unstable scenarios). ``You never need to move that quickly or snap to a position. We don't always want the robot to move fast'' Said he.

(4) Efficiency. P7, P9, P14 simultaneously came up with the idea of ``dynamic scale''. Noticing people need different speeds for different tasks, they all suggested scaling up for efficiency when precision is not needed, and being able to quickly toggle between scaling up and down. P7 gave an example, ``Set a larger scalar during translation (efficiency), and make it smaller for grasp (precision).''

\subsubsection{Mirror -- Reverse the Embodied Motion.}
Users had different preferences over motion reversal on one axis in ``Mirror'' mode. When we designed this feature to have more comfortable ranges of hand motion at the cost of intuitiveness to some extent, 
we were excited to find that P4 had the instinct to move forward when she wanted the robot to come back. She thought the reversed motion felt more natural to her. She related this to her habit of using a trackpad, ``I always used similar mode for scroll wheel or trackpad motion, like swipe up to move a page down. Others may prefer swiping down to move it down.'' This appreciation of our design acknowledges the utility of accommodating various interaction habits in populations. This accommodation might have a profound impact on the perceived usefulness of interaction techniques, especially when they are first introduced.  % implication

% P16 commented, ``Mirror mode is also necessary because the mode I use depends on where I stand''. She preferred mirrored motion when the target object was between her and the robot so that she could keep her hand comfortable.

That being said, the ``Mirror'' mode remained counter-intuitive for most participants. As P9 said, ``Getting close to the robot is weird especially because you might actually touch it''. For example, at proximity, a moving robot arm might hit a participant's arm, or a closing gripper might accidentally squeeze a participant's hand. Despite the robot safety measures we had deployed in the study, participants still wanted to avoid these scenarios for safety concerns. P9 associated the proximity with potential danger, thus bypassing this feature. P10 felt counterintuitive to have the robot moving in the opposite direction with his hand along the forward-backward axis, and further justified his perception by stating that the mirror mode contradicted everyday tools he used such as a unicycle where rotating wheel forward always moves the unicycle forward not backward. In this example, his experience with everyday tools overcame the analogy (i.e., movement in a mirror) we intended to have, leading to negative perceptions of the ``Mirror'' mode.
% Like P10 said, most people had the intuition in embodiment that ``If I see an object I want to grab, my instinct is to go forward, not backward.''

\subsection{RQ3 Findings: How do we explain the significant differences between freehand and controller-based Arm Robot?}

We found that the controller-based version was overall easier to learn (p-value = 0.012), more comfortable (p-value = 0.00027), and more efficient (p-value = 0.018). We were curious about what has led to these significant differences and investigated with semi-structured interviews where we asked participants to elaborate on their ratings.

\subsubsection{Ease of Learning}
Despite all participants agreeing ``Gripper is hand'' is intuitive, participants who reported that freehand interactions were harder due to: 

(1) Contradiction with real-life experience. P3 commented about different ways to pick things up, ``Normally when I grab something, I don't grab it from the top. I grab from the side.'' In our study, most participants opted to grab the object from the top, allowing a longer distance between the robot and the underlying table for collision avoidance. To grab the object from the top, participants would perform a hand pose pointing downwards which was the counterintuitive scenario mentioned by P3. Despite the fact that the controller-based version also had the gripper grabbing the object from the top, the hand pose needed is same as the one used hand controllers in a natural gameplay setting. This difference of discrepancy between the hand poses in Arm Robot and real-life experience contributed to P3's higher ratings of the controller-based version.

(2) Scare of Body Embodiment with Robot. P12 expressed scare when imagining his hand as the gripper, ``If my hand is a direct mapping of the robotic gripper, I feel it is scary when it is doing something I don't intend.'' From this case, we inferred a correlation between the level of embodiment and the sense of concern when unexpected robot movement happened -- which might be due to latency or sensor jitters.

\subsubsection{Comfort and Efficiency}
The difference between freehand and controller-based methods is beyond how much body embodiment is involved. The effective ranges of motion required from a participant's hand in the two methods are different. In the controller-based version, there is no embodied mapping between the wrist pose and the gripper pose so we picked the most common one in literature which is to have gripper pointing downwards mapped to a hand pose grasping a controller pointing forward (i.e., botton's side up). But in the freehand version, the gripper faced down only when the hand also faced down. Having this hand pose as the neutral orientation might result in later awkward hand orientations if a participant was facing the robot arm with the task in between and wanting the robot to point towards themselves. Though a participant can quickly improve this mapping by freezing the embodiment and adjusting to a position at which participants are facing generally the same direction as the robot arm, or by enabling the mirror mode, the brief moments of awareness affected negatively on both the perceived comfort and efficiency of the freehand version.

While having the hand and gripper aligned in orientation for the sense of embodiment, the root cause of the weird hand pose in the freehand Arm Robot was tracking issues and the difference in ranges of motion. 

(1) Tracking issues. To keep the hand in the FoV of the headset camera, users may have to raise their arms with their hands facing down, which is an uncommon arm gesture in daily life. As P4 mentioned, ``Hand positions are very different than what is natural. Quite tiring.'' Several participants reported mild soreness after the study. Participants were aware that they could request a break anytime.
In contrast, the controller method had perfect tracking and a more comfortable wrist orientation. P2 said, ``I need to remind myself that I need to hold my finger down, but I don't need to raise the controller as high.''

(2) The mismatch in the robot's and hand's ranges of motion. If the robot yields to hand, the robot may pose weirdly and cannot grab things stably. Vice versa, when we adapt hand poses for robot performance, we may sacrifice comfort. As a compromised strategy, we manually added a small rotation offset (~20 degrees) for users who felt sore so they could have a more relaxing hand pose without losing too much sense of embodiment.

(3) Tactile feedback. This factor was for a higher comfort score in controller-based Arm Robot. P14 reported that ``holding a non-existent grip with just your hand is odd compared to actually holding a physical grip via the controller''. P1 and P10 said they were not used to keeping the wrist still while holding hands at certain poses. Having the controller to grasp provided a tangible substrate on which the hand could rest and thus was less tiring to perform than the freehand-version.

\subsection{RQ3 Findings: Did users prefer freehand or controller-based Arm Robot?}

Even though the controller-based Arm Robot outperformed the freehand Arm Robot in multiple metrics as shwon in Fig \ref{fig:eval_result} and Fig \ref{fig:likert_scale}, 12 out of 18 participants preferred the freehand Arm Robot to the controller-based Arm Robot when they were asked during the semi-structure interviews. We were curious what led to their preferences and collected their answers as below:

(1) Most participants appreciated the intuitiveness of embodiment. Instead, controller usage requires a ramp-up period. P8 said, ``For a controller, you need to learn the buttons but the hand is intuitive.''

(2) Some users were once confused about the correspondence of orientations between the controller and the gripper. P3 said, ``Without embodiment on the controller, it is at first hard to figure out orientation.'' Though the mapping of a controller pointing forward to a gripper pointing downward is widely used on existing systems in literature and commercial apps, it first appeared counterintuitive to participants.

(3) Participants also appreciated the extensive sense of presence brought by embodiment. P11 expressed the difference, ``When using the hands, it feels more like I am actually participating in the grabbing motion, but the controller is just like I am using an agent to control it.'' P9 gave more details on why she weighted sense of presence over comfort, ``Even though the controller is more comfortable, I prefer hand. For the hands, you feel more involved. I feel like I am actually doing the task, rather than simply controlling a robot. When grabbing or pinching, the robot is also doing that with its hand.''

% design implications P5 - User suggests UI features to add like example hand gestures etc in controller "5.  Since the controller is more precise, it has more inertia and often overshoot the target"

% how the extent of body embodiment affects usability (qualitative)

% \subsubsection{Should We Enable Rotation Offset for freehand Arm Robot?}
% Without rotation offset, the gripper pose always matched the wrist pose. What if we allow users to modify rotation offset? With rotation offset, the gripper and wrist may point in different directions. Although users appreciated the comfort enabled by a rotation offset in certain situations (\eg gripper pointed downwards to grasp things when the wrist could comfortably face forwards), they said it was counter-intuitive to embody rotation with an obvious offset. We tested this in the pilot study, and U2 said, ``I would say you could have an XYZ offset, like a spatial offset, but the rotation should always match my hand... Yeah, it just felt like it wasn't responding to my rotation the way I wanted to.'' U1 also mentioned the lack of a mental model for controlling something with rotation offset. We suggest that this is because people expect tighter embodiment when controlling with a free hand, but an obvious rotation offset contradicts the sense of embodiment.

% \section{Example Applications}
% \label{app}
% \subsection{Augment Human Capability for Accessibility}
% % larger range of motion, larger force

% \subsection{Record and Replay}

% \subsection{Imitation Learning}

\section{Discussion} \label{implications}

%% Always include digital twin when there is system latency according to our observation in the study
According to user feedback from our study, future designers should incorporate a \textbf{predictive path model for real-time embodied control} when there is a delay between the predicted path and the robotic arm's movement. Without predictive paths, user inputs would be sent directly to the robotic arm, leaving users uncertain about the arm's final position, which can lead to safety risks, reduced accuracy, and decreased usability. In our study, we used a zero-delay digital twin of the physical robot arm, with a maximum line velocity of \SI{2}{m/s} and a line acceleration restricted to \SI{0.2}{m/s^2}. Many users found the digital twin beneficial; for example, P16 noted that it significantly improved their task accuracy, as they could better understand the arm's movements. As such,  visual feedback should include predictive paths with latency to address these concerns.

% According to user feedback during our study, it is important that future designers employ a \textbf{predictive path model for real-time embodied control} when there is a time delay between the predicted path and the arm itself. Without this predictive path, user inputs would be sent directly to the robotic arm without the user knowing where the arm would complete their command. This could lead to safety concerns, a decrease in accuracy, and a decrease in overall usability. In this study, we employed a zero-delay digital twin of the physical robot arm that has maximum line velocity at \SI{2}{m/s} and a restricted line acceleration at \SI{0.2}{m/s^2}. Many users found the twin useful. P16 felt that they would not be able to tell how or where the arm was moving without it and mentioned a significant improvement in their accuracy at carrying out tasks. As such, visual feedback in such settings need to include predictive paths with latency to alleviate the above concerns. 

% Always include the feature of Freeze/Unfreeze
Another key feature was \textbf{the ability to freeze/unfreeze the robot at any point during operation}. Users identified this feature essential for both safety and relocation. For safety, it allows users to quickly disable the robot to prevent unwanted collisions, particularly when the user is within the robot's range of motion. For relocation, the freeze/unfreeze function compensates for the mismatch between human and robotic arm capabilities, such as limited reach or wrist rotation. By freezing the robot at their limit, repositioning themselves, and then unfreezing, users could effectively extend their control, utilizing the robot’s full range of motion.

% Another important feature was \textbf{the ability to freeze/unfreeze the robot at any point of operation}. Users felt that this feature was necessary for both safety and relocation. Regarding safety, in the event that the user was within the robot's range of motion, the ability to quickly disable the robot to prevent unwanted collisions was paramount to the user's safety. As for relocation, the range of motion for humans does not match that of the robotic arm. For example, a person's arm can only reach a certain height or a person's wrist can only rotate to a certain degree. As such, the limited user input would result in limited robot range of motion. However, by using the freeze/unfreeze feature, users were able to freeze the robot at their limit, relocate and reposition, unfreeze, and continue their task, thereby achieving the robot's full range of motion. 

Future designers should consider incorporating \textbf{the ability to personalize spatial mapping}, including features like scaling and mirroring. While not universally preferred, these modes proved useful in certain contexts. For instance, scaling mode allows users to adjust the robot arm’s movement relative to their own, enabling larger movements with less effort when scaled up and more precise control when scaled down. An enhancement could involve dynamic scaling that increases during translation and decreases during grasping, optimizing for efficiency, comfort, and precision.
Regarding mirror mode, while most participants favored the default non-mirrored setting, one participant found mirror mode to be the most intuitive. Further inquiry revealed a link between their preference for trackpad scrolling direction and their preferred method of robot control, suggesting a correlation between motion reversal preferences in different interfaces.

% Future designers should also include \textbf{the ability to personalize spatial mapping}. Our study specifically explored scaling and mirroring. Although these modes were not preferred by all users, they still provided useful tools in certain circumstances. For example, scaling mode, which allows users to increase or decrease the robot arm's movement relative to their movement, offered users the option to make large movements with less physical effort when scaled up, but also offered the option to make more precise movements when scaled down. An improvement to this could entail a dynamic scaling system that increases up during translation and decreases during grasp to balance between efficiency, comfort, and precision.
% Regarding mirror mode, while most of our participants preferred the default non-mirrored mode, one participant treated mirror mode as most intuitive. Upon further inquiry, we found an interesting link between their preference in computer trackpad scrolling direction and their preferred method of robot control on whether there is motion reversal. 

Future personalization should include \textbf{hand-gripper correspondence editing}, enabling users to map individual points on their arms and hands to the robot arm, as mobility varies by age, health, and habits. Since the robot arm orients itself to the user’s virtual gripper displayed on their hand, we allowed users to rotate the virtual gripper to match their range of motion. This adjustment helped reduce strain and discomfort during tasks. For example, in a pick-and-place task requiring users to point directly at the tabletop, raising the elbow to an awkward angle caused discomfort over time. By rotating the virtual gripper to a more comfortable position, users could perform the task without strain.

% Future personalization should consider \textbf{hand-gripper correspondence editing}, allowing users to map individual points on their arms and hands to the controlled robot arm. This is because individuals because their mobility may vary by age, health, habits \etc. Given that the robot arm will always orient itself to the user's virtual gripper displayed on their hand, we gave users the option to rotate the virtual gripper and personalize their settings given the users' differing ranges of motion. By doing so, users could reduce strain and discomfort while performing certain tasks. For example, our pick and place task required users to point their hand directly towards the tabletop, thereby requiring users to raise their elbow to an awkward angle, causing discomfort over time. By employing the rotation function to reach a comfortable position, users were able to carry out the task easily without experiencing discomfort.

Finally, future designs should \textbf{expand feedback modalities beyond visuals} to enhance control of the robot arm. Some participantsnoted challenges with depth perception and the absence of force feedback. To address these concerns, we suggest exploring enhanced visual and physical feedback. Enhanced visual feedback could include 
external cameras for more accurate spatial representation or improved user interfaces with wearable sensors for better depth indication. For physical feedback, P12 suggested that a haptic feedback glove could simulate the gripping sensation when the claw grasps an object, thereby improving the teleoperation experience.  

%anything from adding external cameras in environments to produce a more accurate depiction of the physical space to employing an improved user interface and wearable sensing within the headset to indicate depth. For physical feedback, P12 suggested that a haptic feedback glove may be useful to allow the user to experience a gripping sensation when the claw has gripped the object, 
% \input{chapters/8_limitation}
\section{Conclusion}
\label{conclusion}

In this paper, we introduce Arm Robot, an AR-enhanced robot arm teleoperation system designed to overcome HCI challenges in understanding human-robot correspondence, inaccurate perception and limited action space.
By enabling users to adjust the spatial mapping between their movements and the robot's actions, and providing real-time visual feedback on the robot's capabilities, Arm Robot significantly improves control experience for non-technical users.
Specifically, we incorporate several features of ``Freeze / Unfreeze'', ``Scale,'' and ``Mirror'' to allow users to change perspectives and expand action space, along with real-time visualization for feedback in human-robot discrepancies.
Our findings from a user study with 18 participants reveal that Arm Robot effectively mitigate challenges posed by spatial and temporal discrepancies between humans and robots. 
Moreover, our study highlighted the importance of embodiment in human-robot interaction, demonstrating that the degree of embodiment significantly influences user perception and usability. We hope our work can inspire future advances in embodied interactions within human-robot interaction, aiming to make robot control more accessible and intuitive for a broader range of users.

\balance
\bibliographystyle{ACM-Reference-Format}

% \bibliography{chapters/main}

\begin{thebibliography}{48}

%%% the Web address.

\ifx \showCODEN    \undefined \def \showCODEN     #1{\unskip}     \fi
\ifx \showDOI      \undefined \def \showDOI       #1{#1}\fi
\ifx \showISBNx    \undefined \def \showISBNx     #1{\unskip}     \fi
\ifx \showISBNxiii \undefined \def \showISBNxiii  #1{\unskip}     \fi
\ifx \showISSN     \undefined \def \showISSN      #1{\unskip}     \fi
\ifx \showLCCN     \undefined \def \showLCCN      #1{\unskip}     \fi
\ifx \shownote     \undefined \def \shownote      #1{#1}          \fi
\ifx \showarticletitle \undefined \def \showarticletitle #1{#1}   \fi
\ifx \showURL      \undefined \def \showURL       {\relax}        \fi
% invisible to TeX
\providecommand\bibfield[2]{#2}
\providecommand\bibinfo[2]{#2}
\providecommand\natexlab[1]{#1}
\providecommand\showeprint[2][]{arXiv:#2}

\bibitem[Abtahi et~al\mbox{.}(2022)]%
        {abtahi_beyond_2022}
\bibfield{author}{\bibinfo{person}{Parastoo Abtahi}, \bibinfo{person}{Sidney~Q. Hough}, \bibinfo{person}{James~A. Landay}, {and} \bibinfo{person}{Sean Follmer}.} \bibinfo{year}{2022}\natexlab{}.
\newblock \showarticletitle{{Beyond {Being} {Real}: {A} {Sensorimotor} {Control} {Perspective} on {Interactions} in {Virtual} {Reality}}}. In \bibinfo{booktitle}{\emph{{CHI} {Conference} on {Human} {Factors} in {Computing} {Systems}}}. \bibinfo{publisher}{ACM}, \bibinfo{address}{New Orleans LA USA}, \bibinfo{pages}{1--17}.
\newblock
\showISBNx{978-1-4503-9157-3}
\urldef\tempurl%
\url{https://doi.org/10.1145/3491102.3517706}
\showDOI{\tempurl}


\bibitem[Antle et~al\mbox{.}(2011)]%
        {antle2011embodied}
\bibfield{author}{\bibinfo{person}{Alissa~N Antle}, \bibinfo{person}{Paul Marshall}, {and} \bibinfo{person}{Elise van~den Hoven}.} \bibinfo{year}{2011}\natexlab{}.
\newblock \showarticletitle{{Embodied Interaction: Theory and Practice in HCI}}. In \bibinfo{booktitle}{\emph{Proc. CHI}}, Vol.~\bibinfo{volume}{11}.
\newblock
\urldef\tempurl%
\url{https://doi.org/10.1145/1979742.1979592}
\showDOI{\tempurl}


\bibitem[Arunachalam et~al\mbox{.}(2022)]%
        {arunachalam_holo-dex_2022}
\bibfield{author}{\bibinfo{person}{Sridhar~Pandian Arunachalam}, \bibinfo{person}{Irmak Güzey}, \bibinfo{person}{Soumith Chintala}, {and} \bibinfo{person}{Lerrel Pinto}.} \bibinfo{year}{2022}\natexlab{}.
\newblock \bibinfo{title}{{Holo-{Dex}: {Teaching} {Dexterity} With {Immersive} {Mixed} {Reality}}}.
\newblock
\newblock
\urldef\tempurl%
\url{http://arxiv.org/abs/2210.06463}
\showURL{%
\tempurl}
\newblock
\shownote{arXiv:2210.06463 [cs]}.


\bibitem[Bambuŝek et~al\mbox{.}(2019)]%
        {bambusek_combining_2019}
\bibfield{author}{\bibinfo{person}{Daniel Bambuŝek}, \bibinfo{person}{Zdeněk Materna}, \bibinfo{person}{Michal Kapinus}, \bibinfo{person}{Vítězslav Beran}, {and} \bibinfo{person}{Pavel Smrž}.} \bibinfo{year}{2019}\natexlab{}.
\newblock \showarticletitle{{Combining {Interactive} {Spatial} {Augmented} {Reality} With {Head}-{Mounted} {Display} for {End}-{User} {Collaborative} {Robot} {Programming}}}. In \bibinfo{booktitle}{\emph{2019 28th {IEEE} {International} {Conference} on {Robot} and {Human} {Interactive} {Communication} ({RO}-{MAN})}}. \bibinfo{pages}{1--8}.
\newblock
\urldef\tempurl%
\url{https://doi.org/10.1109/RO-MAN46459.2019.8956315}
\showDOI{\tempurl}
\newblock
\shownote{ISSN: 1944-9437}.


\bibitem[Blackett et~al\mbox{.}(2022)]%
        {blackett_effects_2022}
\bibfield{author}{\bibinfo{person}{Claire Blackett}, \bibinfo{person}{Alexandra Fernandes}, \bibinfo{person}{Espen Teigen}, {and} \bibinfo{person}{Thomas Thoresen}.} \bibinfo{year}{2022}\natexlab{}.
\newblock \showarticletitle{{Effects of {Signal} {Latency} on {Human} {Performance} in {Teleoperations}}}. In \bibinfo{booktitle}{\emph{Human {Interaction}, {Emerging} {Technologies} and {Future} {Systems} {V}}}, \bibfield{editor}{\bibinfo{person}{Tareq Ahram} {and} \bibinfo{person}{Redha Taiar}} (Eds.). \bibinfo{publisher}{Springer International Publishing}, \bibinfo{address}{Cham}, \bibinfo{pages}{386--393}.
\newblock
\showISBNx{978-3-030-85540-6}
\urldef\tempurl%
\url{https://doi.org/10.1007/978-3-030-85540-\_50}
\showDOI{\tempurl}


\bibitem[Cordeil et~al\mbox{.}(2020)]%
        {cordeil_embodied_2020}
\bibfield{author}{\bibinfo{person}{Maxime Cordeil}, \bibinfo{person}{Benjamin Bach}, \bibinfo{person}{Andrew Cunningham}, \bibinfo{person}{Bastian Montoya}, \bibinfo{person}{Ross~T. Smith}, \bibinfo{person}{Bruce~H. Thomas}, {and} \bibinfo{person}{Tim Dwyer}.} \bibinfo{year}{2020}\natexlab{}.
\newblock \showarticletitle{{Embodied {Axes}: {Tangible}, {Actuated} {Interaction} for {3D} {Augmented} {Reality} {Data} {Spaces}}}. In \bibinfo{booktitle}{\emph{Proceedings of the 2020 {CHI} {Conference} on {Human} {Factors} in {Computing} {Systems}}}. \bibinfo{publisher}{ACM}, \bibinfo{address}{Honolulu HI USA}, \bibinfo{pages}{1--12}.
\newblock
\showISBNx{978-1-4503-6708-0}
\urldef\tempurl%
\url{https://doi.org/10.1145/3313831.3376613}
\showDOI{\tempurl}


\bibitem[Cutsuridis et~al\mbox{.}(2011)]%
        {cutsuridis2011perception}
\bibfield{author}{\bibinfo{person}{Vassilis Cutsuridis}, \bibinfo{person}{Amir Hussain}, {and} \bibinfo{person}{John~G Taylor}.} \bibinfo{year}{2011}\natexlab{}.
\newblock \bibinfo{booktitle}{\emph{Perception-action cycle: Models, architectures, and hardware}}.
\newblock \bibinfo{publisher}{Springer Science \& Business Media}.
\newblock


\bibitem[DelPreto et~al\mbox{.}(2020)]%
        {delpreto_helping_2020}
\bibfield{author}{\bibinfo{person}{Joseph DelPreto}, \bibinfo{person}{Jeffrey~I. Lipton}, \bibinfo{person}{Lindsay Sanneman}, \bibinfo{person}{Aidan~J. Fay}, \bibinfo{person}{Christopher Fourie}, \bibinfo{person}{Changhyun Choi}, {and} \bibinfo{person}{Daniela Rus}.} \bibinfo{year}{2020}\natexlab{}.
\newblock \showarticletitle{{Helping {Robots} {Learn}: {A} {Human}-{Robot} {Master}-{Apprentice} {Model} {Using} {Demonstrations} via {Virtual} {Reality} {Teleoperation}}}. In \bibinfo{booktitle}{\emph{2020 {IEEE} {International} {Conference} on {Robotics} and {Automation} ({ICRA})}}. \bibinfo{pages}{10226--10233}.
\newblock
\urldef\tempurl%
\url{https://doi.org/10.1109/ICRA40945.2020.9196754}
\showDOI{\tempurl}
\newblock
\shownote{ISSN: 2577-087X}.


\bibitem[Dom{\'\i}nguez-Vidal et~al\mbox{.}(2022)]%
        {dominguez2022perception}
\bibfield{author}{\bibinfo{person}{Jose~Enrique Dom{\'\i}nguez-Vidal}, \bibinfo{person}{Nicolas Rodriguez}, \bibinfo{person}{Rene Alquezar}, {and} \bibinfo{person}{Alberto Sanfeliu}.} \bibinfo{year}{2022}\natexlab{}.
\newblock \showarticletitle{Perception-Intention-Action Cycle in Human-Robot Collaborative Tasks}.
\newblock \bibinfo{journal}{\emph{arXiv preprint arXiv:2206.00304}} (\bibinfo{year}{2022}).
\newblock


\bibitem[Dourish(2001)]%
        {dourish_where_2001}
\bibfield{author}{\bibinfo{person}{Paul Dourish}.} \bibinfo{year}{2001}\natexlab{}.
\newblock \bibinfo{booktitle}{\emph{{Where the {Action} Is: {The} {Foundations} of {Embodied} {Interaction}}}}.
\newblock \bibinfo{publisher}{MIT Press}.
\newblock
\showISBNx{978-0-262-54178-7}
\newblock
\shownote{Google-Books-ID: DCIy2zxrCqcC}.


\bibitem[Duan et~al\mbox{.}(2023)]%
        {duan_ar2-d2training_2023}
\bibfield{author}{\bibinfo{person}{Jiafei Duan}, \bibinfo{person}{Yi~Ru Wang}, \bibinfo{person}{Mohit Shridhar}, \bibinfo{person}{Dieter Fox}, {and} \bibinfo{person}{Ranjay Krishna}.} \bibinfo{year}{2023}\natexlab{}.
\newblock \bibinfo{title}{{AR2}-{D2}:{Training} a {Robot} {Without} a {Robot}}.
\newblock
\newblock
\urldef\tempurl%
\url{http://arxiv.org/abs/2306.13818}
\showURL{%
\tempurl}
\newblock
\shownote{arXiv:2306.13818 [cs]}.


\bibitem[Dybvik et~al\mbox{.}(2021)]%
        {dybvik_low-cost_2021}
\bibfield{author}{\bibinfo{person}{Henrikke Dybvik}, \bibinfo{person}{Martin Løland}, \bibinfo{person}{Achim Gerstenberg}, \bibinfo{person}{Kristoffer~Bjørnerud Slåttsveen}, {and} \bibinfo{person}{Martin Steinert}.} \bibinfo{year}{2021}\natexlab{}.
\newblock \showarticletitle{{A Low-Cost Predictive Display for Teleoperation: {Investigating} Effects on Human Performance and Workload}}.
\newblock \bibinfo{journal}{\emph{International Journal of Human-Computer Studies}}  \bibinfo{volume}{145} (\bibinfo{year}{2021}), \bibinfo{pages}{102536}.
\newblock
\showISSN{1071-5819}
\urldef\tempurl%
\url{https://doi.org/10.1016/j.ijhcs.2020.102536}
\showDOI{\tempurl}


\bibitem[Feuchtner and Müller(2017)]%
        {feuchtner_extending_2017}
\bibfield{author}{\bibinfo{person}{Tiare Feuchtner} {and} \bibinfo{person}{Jörg Müller}.} \bibinfo{year}{2017}\natexlab{}.
\newblock \showarticletitle{{Extending the {Body} for {Interaction} With {Reality}}}. In \bibinfo{booktitle}{\emph{Proceedings of the 2017 {CHI} {Conference} on {Human} {Factors} in {Computing} {Systems}}}. \bibinfo{publisher}{ACM}, \bibinfo{address}{Denver Colorado USA}, \bibinfo{pages}{5145--5157}.
\newblock
\showISBNx{978-1-4503-4655-9}
\urldef\tempurl%
\url{https://doi.org/10.1145/3025453.3025689}
\showDOI{\tempurl}


\bibitem[Fritsche et~al\mbox{.}(2015)]%
        {fritsche_first-person_2015}
\bibfield{author}{\bibinfo{person}{Lars Fritsche}, \bibinfo{person}{Felix Unverzag}, \bibinfo{person}{Jan Peters}, {and} \bibinfo{person}{Roberto Calandra}.} \bibinfo{year}{2015}\natexlab{}.
\newblock \showarticletitle{{First-Person Tele-Operation of a Humanoid Robot}}. In \bibinfo{booktitle}{\emph{2015 {IEEE}-{RAS} 15th {International} {Conference} on {Humanoid} {Robots} ({Humanoids})}}. \bibinfo{pages}{997--1002}.
\newblock
\urldef\tempurl%
\url{https://doi.org/10.1109/HUMANOIDS.2015.7363475}
\showDOI{\tempurl}


\bibitem[Gaurav et~al\mbox{.}(2019)]%
        {gaurav_deep_2019}
\bibfield{author}{\bibinfo{person}{Sanket Gaurav}, \bibinfo{person}{Zainab Al-Qurashi}, \bibinfo{person}{Amey Barapatre}, \bibinfo{person}{George Maratos}, \bibinfo{person}{Tejas Sarma}, {and} \bibinfo{person}{Brian~D. Ziebart}.} \bibinfo{year}{2019}\natexlab{}.
\newblock \showarticletitle{{Deep {Correspondence} {Learning} for {Effective} {Robotic} {Teleoperation} Using {Virtual} {Reality}}}. In \bibinfo{booktitle}{\emph{2019 {IEEE}-{RAS} 19th {International} {Conference} on {Humanoid} {Robots} ({Humanoids})}}. \bibinfo{pages}{477--483}.
\newblock
\urldef\tempurl%
\url{https://doi.org/10.1109/Humanoids43949.2019.9035031}
\showDOI{\tempurl}
\newblock
\shownote{ISSN: 2164-0580}.


\bibitem[Geurs and Van~Wee(2004)]%
        {geurs2004accessibility}
\bibfield{author}{\bibinfo{person}{Karst~T Geurs} {and} \bibinfo{person}{Bert Van~Wee}.} \bibinfo{year}{2004}\natexlab{}.
\newblock \showarticletitle{Accessibility evaluation of land-use and transport strategies: review and research directions}.
\newblock \bibinfo{journal}{\emph{Journal of Transport geography}} \bibinfo{volume}{12}, \bibinfo{number}{2} (\bibinfo{year}{2004}), \bibinfo{pages}{127--140}.
\newblock


\bibitem[Kaarlela et~al\mbox{.}(2022)]%
        {kaarlela2022digital}
\bibfield{author}{\bibinfo{person}{Tero Kaarlela}, \bibinfo{person}{Paulo Padrao}, \bibinfo{person}{Tomi Pitk{\"a}aho}, \bibinfo{person}{Sakari Piesk{\"a}}, {and} \bibinfo{person}{Leonardo Bobadilla}.} \bibinfo{year}{2022}\natexlab{}.
\newblock \showarticletitle{Digital twins utilizing XR-technology as robotic training tools}.
\newblock \bibinfo{journal}{\emph{Machines}} \bibinfo{volume}{11}, \bibinfo{number}{1} (\bibinfo{year}{2022}), \bibinfo{pages}{13}.
\newblock


\bibitem[Khundam et~al\mbox{.}(2021)]%
        {khundam2021comparative}
\bibfield{author}{\bibinfo{person}{Chaowanan Khundam}, \bibinfo{person}{Varunyu Vorachart}, \bibinfo{person}{Patibut Preeyawongsakul}, \bibinfo{person}{Witthaya Hosap}, {and} \bibinfo{person}{Fr{\'e}d{\'e}ric No{\"e}l}.} \bibinfo{year}{2021}\natexlab{}.
\newblock \showarticletitle{A comparative study of interaction time and usability of using controllers and hand tracking in virtual reality training}. In \bibinfo{booktitle}{\emph{Informatics}}, Vol.~\bibinfo{volume}{8}. MDPI, \bibinfo{pages}{60}.
\newblock


\bibitem[LaViola~Jr et~al\mbox{.}(2017)]%
        {laviola20173d}
\bibfield{author}{\bibinfo{person}{Joseph~J LaViola~Jr}, \bibinfo{person}{Ernst Kruijff}, \bibinfo{person}{Ryan~P McMahan}, \bibinfo{person}{Doug Bowman}, {and} \bibinfo{person}{Ivan~P Poupyrev}.} \bibinfo{year}{2017}\natexlab{}.
\newblock \bibinfo{booktitle}{\emph{{3D User Interfaces: Theory and Practice}}}.
\newblock \bibinfo{publisher}{Addison-Wesley Professional}.
\newblock


\bibitem[Li et~al\mbox{.}(2020)]%
        {li_mobile_2020}
\bibfield{author}{\bibinfo{person}{Shuang Li}, \bibinfo{person}{Jiaxi Jiang}, \bibinfo{person}{Philipp Ruppel}, \bibinfo{person}{Hongzhuo Liang}, \bibinfo{person}{Xiaojian Ma}, \bibinfo{person}{Norman Hendrich}, \bibinfo{person}{Fuchun Sun}, {and} \bibinfo{person}{Jianwei Zhang}.} \bibinfo{year}{2020}\natexlab{}.
\newblock \showarticletitle{{A {Mobile} {Robot} {Hand}-{Arm} {Teleoperation} {System} by {Vision} and {IMU}}}. In \bibinfo{booktitle}{\emph{2020 {IEEE}/{RSJ} {International} {Conference} on {Intelligent} {Robots} and {Systems} ({IROS})}}. \bibinfo{pages}{10900--10906}.
\newblock
\urldef\tempurl%
\url{https://doi.org/10.1109/IROS45743.2020.9340738}
\showDOI{\tempurl}
\newblock
\shownote{ISSN: 2153-0866}.


\bibitem[Lum et~al\mbox{.}(2009)]%
        {lum_teleoperation_2009}
\bibfield{author}{\bibinfo{person}{Mitchell J.~H. Lum}, \bibinfo{person}{Jacob Rosen}, \bibinfo{person}{Hawkeye King}, \bibinfo{person}{Diana C.~W. Friedman}, \bibinfo{person}{Thomas~S. Lendvay}, \bibinfo{person}{Andrew~S. Wright}, \bibinfo{person}{Mika~N. Sinanan}, {and} \bibinfo{person}{Blake Hannaford}.} \bibinfo{year}{2009}\natexlab{}.
\newblock \showarticletitle{{Teleoperation in Surgical Robotics-Network Latency Effects on Surgical Performance}}.
\newblock \bibinfo{journal}{\emph{Annual International Conference of the IEEE Engineering in Medicine and Biology Society. IEEE Engineering in Medicine and Biology Society. Annual International Conference}}  \bibinfo{volume}{2009} (\bibinfo{year}{2009}), \bibinfo{pages}{6860--6863}.
\newblock
\showISSN{2375-7477}
\urldef\tempurl%
\url{https://doi.org/10.1109/IEMBS.2009.5333120}
\showDOI{\tempurl}


\bibitem[Matsas and Vosniakos(2017)]%
        {matsas_design_2017}
\bibfield{author}{\bibinfo{person}{Elias Matsas} {and} \bibinfo{person}{George-Christopher Vosniakos}.} \bibinfo{year}{2017}\natexlab{}.
\newblock \showarticletitle{{Design of a Virtual Reality Training System for Human–robot Collaboration in Manufacturing Tasks}}.
\newblock \bibinfo{journal}{\emph{International Journal on Interactive Design and Manufacturing (IJIDeM)}} \bibinfo{volume}{11}, \bibinfo{number}{2} (\bibinfo{year}{2017}), \bibinfo{pages}{139--153}.
\newblock
\showISSN{1955-2505}
\urldef\tempurl%
\url{https://doi.org/10.1007/s12008-015-0259-2}
\showDOI{\tempurl}


\bibitem[Pei et~al\mbox{.}(2023)]%
        {pei_embodied_2023}
\bibfield{author}{\bibinfo{person}{Siyou Pei}, \bibinfo{person}{Alexander Chen}, \bibinfo{person}{Chen Chen}, \bibinfo{person}{Franklin~Mingzhe Li}, \bibinfo{person}{Megan Fozzard}, \bibinfo{person}{Hao-Yun Chi}, \bibinfo{person}{Nadir Weibel}, \bibinfo{person}{Patrick Carrington}, {and} \bibinfo{person}{Yang Zhang}.} \bibinfo{year}{2023}\natexlab{}.
\newblock \showarticletitle{{Embodied {Exploration}: {Facilitating} {Remote} {Accessibility} {Assessment} for {Wheelchair} {Users} With {Virtual} {Reality}}}. In \bibinfo{booktitle}{\emph{Proceedings of the 25th {International} {ACM} {SIGACCESS} {Conference} on {Computers} and {Accessibility}}} \emph{(\bibinfo{series}{{ASSETS} '23})}. \bibinfo{publisher}{Association for Computing Machinery}, \bibinfo{address}{New York, NY, USA}, \bibinfo{pages}{1--17}.
\newblock
\showISBNx{9798400702204}
\urldef\tempurl%
\url{https://doi.org/10.1145/3597638.3608410}
\showDOI{\tempurl}


\bibitem[Pei et~al\mbox{.}(2022)]%
        {pei_hand_2022}
\bibfield{author}{\bibinfo{person}{Siyou Pei}, \bibinfo{person}{Alexander Chen}, \bibinfo{person}{Jaewook Lee}, {and} \bibinfo{person}{Yang Zhang}.} \bibinfo{year}{2022}\natexlab{}.
\newblock \showarticletitle{{Hand {Interfaces}: {Using} {Hands} to {Imitate} {Objects} in {AR}/{VR} for {Expressive} {Interactions}}}. In \bibinfo{booktitle}{\emph{{CHI} {Conference} on {Human} {Factors} in {Computing} {Systems}}}. \bibinfo{publisher}{ACM}, \bibinfo{address}{New Orleans LA USA}, \bibinfo{pages}{1--16}.
\newblock
\showISBNx{978-1-4503-9157-3}
\urldef\tempurl%
\url{https://doi.org/10.1145/3491102.3501898}
\showDOI{\tempurl}


\bibitem[Piumsomboon et~al\mbox{.}(2018a)]%
        {piumsomboon_snow_2018}
\bibfield{author}{\bibinfo{person}{Thammathip Piumsomboon}, \bibinfo{person}{Gun~A. Lee}, {and} \bibinfo{person}{Mark Billinghurst}.} \bibinfo{year}{2018}\natexlab{a}.
\newblock \showarticletitle{{Snow {Dome}: {A} {Multi}-{Scale} {Interaction} in {Mixed} {Reality} {Remote} {Collaboration}}}. In \bibinfo{booktitle}{\emph{Extended {Abstracts} of the 2018 {CHI} {Conference} on {Human} {Factors} in {Computing} {Systems}}} \emph{(\bibinfo{series}{{CHI} {EA} '18})}. \bibinfo{publisher}{Association for Computing Machinery}, \bibinfo{address}{New York, NY, USA}, \bibinfo{pages}{1--4}.
\newblock
\showISBNx{978-1-4503-5621-3}
\urldef\tempurl%
\url{https://doi.org/10.1145/3170427.3186495}
\showDOI{\tempurl}


\bibitem[Piumsomboon et~al\mbox{.}(2018b)]%
        {piumsomboon_mini-me_2018}
\bibfield{author}{\bibinfo{person}{Thammathip Piumsomboon}, \bibinfo{person}{Gun~A. Lee}, \bibinfo{person}{Jonathon~D. Hart}, \bibinfo{person}{Barrett Ens}, \bibinfo{person}{Robert~W. Lindeman}, \bibinfo{person}{Bruce~H. Thomas}, {and} \bibinfo{person}{Mark Billinghurst}.} \bibinfo{year}{2018}\natexlab{b}.
\newblock \showarticletitle{{Mini-{Me}: {An} {Adaptive} {Avatar} for {Mixed} {Reality} {Remote} {Collaboration}}}. In \bibinfo{booktitle}{\emph{Proceedings of the 2018 {CHI} {Conference} on {Human} {Factors} in {Computing} {Systems}}} \emph{(\bibinfo{series}{{CHI} '18})}. \bibinfo{publisher}{Association for Computing Machinery}, \bibinfo{address}{New York, NY, USA}, \bibinfo{pages}{1--13}.
\newblock
\showISBNx{978-1-4503-5620-6}
\urldef\tempurl%
\url{https://doi.org/10.1145/3173574.3173620}
\showDOI{\tempurl}


\bibitem[Pot et~al\mbox{.}(2021)]%
        {pot2021perceived}
\bibfield{author}{\bibinfo{person}{Felix~Johan Pot}, \bibinfo{person}{Bert van Wee}, {and} \bibinfo{person}{Taede Tillema}.} \bibinfo{year}{2021}\natexlab{}.
\newblock \showarticletitle{Perceived accessibility: What it is and why it differs from calculated accessibility measures based on spatial data}.
\newblock \bibinfo{journal}{\emph{Journal of Transport Geography}}  \bibinfo{volume}{94} (\bibinfo{year}{2021}), \bibinfo{pages}{103090}.
\newblock


\bibitem[Qin et~al\mbox{.}(2023)]%
        {qin_anyteleop_2023}
\bibfield{author}{\bibinfo{person}{Yuzhe Qin}, \bibinfo{person}{Wei Yang}, \bibinfo{person}{Binghao Huang}, \bibinfo{person}{Karl Van~Wyk}, \bibinfo{person}{Hao Su}, \bibinfo{person}{Xiaolong Wang}, \bibinfo{person}{Yu-Wei Chao}, {and} \bibinfo{person}{Dieter Fox}.} \bibinfo{year}{2023}\natexlab{}.
\newblock \bibinfo{title}{{AnyTeleop}: {A} {General} {Vision}-{Based} {Dexterous} {Robot} {Arm}-{Hand} {Teleoperation} {System}}.
\newblock
\newblock
\urldef\tempurl%
\url{http://arxiv.org/abs/2307.04577}
\showURL{%
\tempurl}
\newblock
\shownote{arXiv:2307.04577 [cs]}.


\bibitem[Radalytica(2020)]%
        {radalytica_precise_2020}
\bibfield{author}{\bibinfo{person}{Radalytica}.} \bibinfo{year}{2020}\natexlab{}.
\newblock \bibinfo{title}{{Precise Real-Time {Universal} Robot Control Using {3D} Move - Pouring a Water}}.
\newblock
\newblock
\urldef\tempurl%
\url{https://www.youtube.com/watch?v=3C3JtE3glnk}
\showURL{%
\tempurl}


\bibitem[Rantamaa et~al\mbox{.}(2023)]%
        {rantamaa2023comparison}
\bibfield{author}{\bibinfo{person}{Hanna-Riikka Rantamaa}, \bibinfo{person}{Jari Kangas}, \bibinfo{person}{Sriram~Kishore Kumar}, \bibinfo{person}{Helena Mehtonen}, \bibinfo{person}{Jorma J{\"a}rnstedt}, {and} \bibinfo{person}{Roope Raisamo}.} \bibinfo{year}{2023}\natexlab{}.
\newblock \showarticletitle{Comparison of a vr stylus with a controller, hand tracking, and a mouse for object manipulation and medical marking tasks in virtual reality}.
\newblock \bibinfo{journal}{\emph{Applied Sciences}} \bibinfo{volume}{13}, \bibinfo{number}{4} (\bibinfo{year}{2023}), \bibinfo{pages}{2251}.
\newblock


\bibitem[Robot(2024)]%
        {universal_robot_collaborative}
\bibfield{author}{\bibinfo{person}{Universal Robot}.} \bibinfo{year}{2024}\natexlab{}.
\newblock \bibinfo{title}{Collaborative robotic automation {\textbar} {Cobots} from {Universal} {Robots}}.
\newblock
\newblock
\urldef\tempurl%
\url{https://www.universal-robots.com/, https://www.universal-robots.com/products/ur5-robot/}
\showURL{%
\tempurl}


\bibitem[Robotics(2024)]%
        {franka_robotics_franka}
\bibfield{author}{\bibinfo{person}{Franka Robotics}.} \bibinfo{year}{2024}\natexlab{}.
\newblock \bibinfo{title}{Franka {Emika} {Robot} {Specifications}}.
\newblock
\newblock
\urldef\tempurl%
\url{https://franka.de/research}
\showURL{%
\tempurl}


\bibitem[Robotics(2022)]%
        {nordbo_robotics_mimic_2022}
\bibfield{author}{\bibinfo{person}{Nordbo Robotics}.} \bibinfo{year}{2022}\natexlab{}.
\newblock \bibinfo{title}{{Mimic With {IR} {Tracker}: {Demonstration} {\textbar} {Nordbo} {Robotics}}}.
\newblock
\newblock
\urldef\tempurl%
\url{https://www.youtube.com/watch?v=WqaS1p9BVHA}
\showURL{%
\tempurl}


\bibitem[Schjerlund et~al\mbox{.}(2021)]%
        {schjerlund_ninja_2021}
\bibfield{author}{\bibinfo{person}{Jonas Schjerlund}, \bibinfo{person}{Kasper Hornbæk}, {and} \bibinfo{person}{Joanna Bergström}.} \bibinfo{year}{2021}\natexlab{}.
\newblock \showarticletitle{{Ninja {Hands}: {Using} {Many} {Hands} to {Improve} {Target} {Selection} in {VR}}}. In \bibinfo{booktitle}{\emph{Proceedings of the 2021 {CHI} {Conference} on {Human} {Factors} in {Computing} {Systems}}}. \bibinfo{publisher}{ACM}, \bibinfo{address}{Yokohama Japan}, \bibinfo{pages}{1--14}.
\newblock
\showISBNx{978-1-4503-8096-6}
\urldef\tempurl%
\url{https://doi.org/10.1145/3411764.3445759}
\showDOI{\tempurl}


\bibitem[Schwartz(2016)]%
        {schwartz2016movement}
\bibfield{author}{\bibinfo{person}{Andrew~B Schwartz}.} \bibinfo{year}{2016}\natexlab{}.
\newblock \showarticletitle{Movement: how the brain communicates with the world}.
\newblock \bibinfo{journal}{\emph{Cell}} \bibinfo{volume}{164}, \bibinfo{number}{6} (\bibinfo{year}{2016}), \bibinfo{pages}{1122--1135}.
\newblock


\bibitem[Sivakumar et~al\mbox{.}(2022)]%
        {sivakumar_robotic_2022}
\bibfield{author}{\bibinfo{person}{Aravind Sivakumar}, \bibinfo{person}{Kenneth Shaw}, {and} \bibinfo{person}{Deepak Pathak}.} \bibinfo{year}{2022}\natexlab{}.
\newblock \bibinfo{title}{{Robotic {Telekinesis}: {Learning} a {Robotic} {Hand} {Imitator} by {Watching} {Humans} on {Youtube}}}.
\newblock
\newblock
\urldef\tempurl%
\url{https://doi.org/10.48550/arXiv.2202.10448}
\showDOI{\tempurl}
\newblock
\shownote{arXiv:2202.10448 [cs]}.


\bibitem[Soares et~al\mbox{.}(2021)]%
        {soares_programming_2021}
\bibfield{author}{\bibinfo{person}{Inês Soares}, \bibinfo{person}{Marcelo Petry}, {and} \bibinfo{person}{António~Paulo Moreira}.} \bibinfo{year}{2021}\natexlab{}.
\newblock \showarticletitle{{Programming {Robots} by {Demonstration} {Using} {Augmented} {Reality}}}.
\newblock \bibinfo{journal}{\emph{Sensors}} \bibinfo{volume}{21}, \bibinfo{number}{17} (\bibinfo{year}{2021}), \bibinfo{pages}{5976}.
\newblock
\showISSN{1424-8220}
\urldef\tempurl%
\url{https://doi.org/10.3390/s21175976}
\showDOI{\tempurl}


\bibitem[Sousa et~al\mbox{.}(2019)]%
        {sousa2019warping}
\bibfield{author}{\bibinfo{person}{Maur{\'\i}cio Sousa}, \bibinfo{person}{Rafael~Kufner dos Anjos}, \bibinfo{person}{Daniel Mendes}, \bibinfo{person}{Mark Billinghurst}, {and} \bibinfo{person}{Joaquim Jorge}.} \bibinfo{year}{2019}\natexlab{}.
\newblock \showarticletitle{{Warping Deixis: Distorting Gestures to Enhance Collaboration}}. In \bibinfo{booktitle}{\emph{Proceedings of the 2019 CHI Conference on Human Factors in Computing Systems}}. \bibinfo{pages}{1--12}.
\newblock
\urldef\tempurl%
\url{https://doi.org/10.1145/3290605.3300838}
\showDOI{\tempurl}


\bibitem[Starke et~al\mbox{.}(2017)]%
        {starke2017evolutionary}
\bibfield{author}{\bibinfo{person}{Sebastian Starke}, \bibinfo{person}{Norman Hendrich}, \bibinfo{person}{Dennis Krupke}, {and} \bibinfo{person}{Jianwei Zhang}.} \bibinfo{year}{2017}\natexlab{}.
\newblock \showarticletitle{{Evolutionary Multi-Objective Inverse Kinematics on Highly Articulated and Humanoid Robots}}. In \bibinfo{booktitle}{\emph{2017 IEEE/RSJ International Conference on Intelligent Robots and Systems (IROS)}}. IEEE, \bibinfo{publisher}{IEEE}, \bibinfo{pages}{6959--6966}.
\newblock
\urldef\tempurl%
\url{https://doi.org/10.1109/IROS.2017.8206620}
\showDOI{\tempurl}


\bibitem[Starke et~al\mbox{.}(2016)]%
        {starke2016efficient}
\bibfield{author}{\bibinfo{person}{Sebastian Starke}, \bibinfo{person}{Norman Hendrich}, \bibinfo{person}{Sven Magg}, {and} \bibinfo{person}{Jianwei Zhang}.} \bibinfo{year}{2016}\natexlab{}.
\newblock \showarticletitle{{An Efficient Hybridization of Genetic Algorithms and Particle Swarm Optimization for Inverse Kinematics}}. In \bibinfo{booktitle}{\emph{2016 IEEE International Conference on Robotics and Biomimetics (ROBIO)}}. IEEE, \bibinfo{publisher}{IEEE}, \bibinfo{pages}{1782--1789}.
\newblock
\urldef\tempurl%
\url{https://doi.org/10.1007/978-3-540-87442-\_96}
\showDOI{\tempurl}


\bibitem[Su et~al\mbox{.}(2022)]%
        {su_mixed-reality-enhanced_2022}
\bibfield{author}{\bibinfo{person}{Yun-Peng Su}, \bibinfo{person}{Xiao-Qi Chen}, \bibinfo{person}{Tony Zhou}, \bibinfo{person}{Christopher Pretty}, {and} \bibinfo{person}{Geoffrey Chase}.} \bibinfo{year}{2022}\natexlab{}.
\newblock \showarticletitle{{Mixed-{Reality}-{Enhanced} {Human}–{Robot} {Interaction} With an {Imitation}-{Based} {Mapping} {Approach} for {Intuitive} {Teleoperation} of a {Robotic} {Arm}-{Hand} {System}}}.
\newblock \bibinfo{journal}{\emph{Applied Sciences}} \bibinfo{volume}{12}, \bibinfo{number}{9} (\bibinfo{year}{2022}), \bibinfo{pages}{4740}.
\newblock
\showISSN{2076-3417}
\urldef\tempurl%
\url{https://doi.org/10.3390/app12094740}
\showDOI{\tempurl}


\bibitem[Suzuki et~al\mbox{.}(2022)]%
        {suzuki_augmented_2022}
\bibfield{author}{\bibinfo{person}{Ryo Suzuki}, \bibinfo{person}{Adnan Karim}, \bibinfo{person}{Tian Xia}, \bibinfo{person}{Hooman Hedayati}, {and} \bibinfo{person}{Nicolai Marquardt}.} \bibinfo{year}{2022}\natexlab{}.
\newblock \showarticletitle{{Augmented {Reality} and {Robotics}: {A} {Survey} and {Taxonomy} for {AR}-Enhanced {Human}-{Robot} {Interaction} and {Robotic} {Interfaces}}}. In \bibinfo{booktitle}{\emph{{CHI} {Conference} on {Human} {Factors} in {Computing} {Systems}}}. \bibinfo{publisher}{ACM}, \bibinfo{address}{New Orleans LA USA}, \bibinfo{pages}{1--33}.
\newblock
\showISBNx{978-1-4503-9157-3}
\urldef\tempurl%
\url{https://doi.org/10.1145/3491102.3517719}
\showDOI{\tempurl}


\bibitem[Tao et~al\mbox{.}(2023)]%
        {tao_embodying_2023}
\bibfield{author}{\bibinfo{person}{Yujie Tao}, \bibinfo{person}{Cheng~Yao Wang}, \bibinfo{person}{Andrew~D Wilson}, \bibinfo{person}{Eyal Ofek}, {and} \bibinfo{person}{Mar Gonzalez-Franco}.} \bibinfo{year}{2023}\natexlab{}.
\newblock \showarticletitle{{Embodying {Physics}-{Aware} {Avatars} in {Virtual} {Reality}}}. In \bibinfo{booktitle}{\emph{Proceedings of the 2023 {CHI} {Conference} on {Human} {Factors} in {Computing} {Systems}}} \emph{(\bibinfo{series}{{CHI} '23})}. \bibinfo{publisher}{Association for Computing Machinery}, \bibinfo{address}{New York, NY, USA}, \bibinfo{pages}{1--15}.
\newblock
\showISBNx{978-1-4503-9421-5}
\urldef\tempurl%
\url{https://doi.org/10.1145/3544548.3580979}
\showDOI{\tempurl}


\bibitem[Togias et~al\mbox{.}(2021)]%
        {togias_virtual_2021}
\bibfield{author}{\bibinfo{person}{Theodoros Togias}, \bibinfo{person}{Christos Gkournelos}, \bibinfo{person}{Panagiotis Angelakis}, \bibinfo{person}{George Michalos}, {and} \bibinfo{person}{Sotiris Makris}.} \bibinfo{year}{2021}\natexlab{}.
\newblock \showarticletitle{{Virtual Reality Environment for Industrial Robot Control and Path Design}}.
\newblock \bibinfo{journal}{\emph{Procedia CIRP}}  \bibinfo{volume}{100} (\bibinfo{year}{2021}), \bibinfo{pages}{133--138}.
\newblock
\showISSN{22128271}
\urldef\tempurl%
\url{https://doi.org/10.1016/j.procir.2021.05.021}
\showDOI{\tempurl}


\bibitem[Wang et~al\mbox{.}(2023)]%
        {wang_explainable_2023}
\bibfield{author}{\bibinfo{person}{Chao Wang}, \bibinfo{person}{Anna Belardinelli}, \bibinfo{person}{Stephan Hasler}, \bibinfo{person}{Theodoros Stouraitis}, \bibinfo{person}{Daniel Tanneberg}, {and} \bibinfo{person}{Michael Gienger}.} \bibinfo{year}{2023}\natexlab{}.
\newblock \showarticletitle{{Explainable {Human}-{Robot} {Training} and {Cooperation} With {Augmented} {Reality}}}. In \bibinfo{booktitle}{\emph{Extended {Abstracts} of the 2023 {CHI} {Conference} on {Human} {Factors} in {Computing} {Systems}}} \emph{(\bibinfo{series}{{CHI} {EA} '23})}. \bibinfo{publisher}{Association for Computing Machinery}, \bibinfo{address}{New York, NY, USA}, \bibinfo{pages}{1--5}.
\newblock
\showISBNx{978-1-4503-9422-2}
\urldef\tempurl%
\url{https://doi.org/10.1145/3544549.3583889}
\showDOI{\tempurl}


\bibitem[Xu et~al\mbox{.}(2018)]%
        {xu_robot_2018}
\bibfield{author}{\bibinfo{person}{Yang Xu}, \bibinfo{person}{Chenguang Yang}, \bibinfo{person}{Junpei Zhong}, \bibinfo{person}{Ning Wang}, {and} \bibinfo{person}{Lijun Zhao}.} \bibinfo{year}{2018}\natexlab{}.
\newblock \showarticletitle{{Robot Teaching by Teleoperation Based on Visual Interaction and Extreme Learning Machine}}.
\newblock \bibinfo{journal}{\emph{Neurocomputing}}  \bibinfo{volume}{275} (\bibinfo{year}{2018}), \bibinfo{pages}{2093--2103}.
\newblock
\showISSN{0925-2312}
\urldef\tempurl%
\url{https://doi.org/10.1016/j.neucom.2017.10.034}
\showDOI{\tempurl}


\bibitem[Yim et~al\mbox{.}(2022)]%
        {yim_wfh-vr_2022}
\bibfield{author}{\bibinfo{person}{Lai~Sum Yim}, \bibinfo{person}{Quang~Tn Vo}, \bibinfo{person}{Ching-I Huang}, \bibinfo{person}{Chi-Ruei Wang}, \bibinfo{person}{Wren McQueary}, \bibinfo{person}{Hsueh-Cheng Wang}, \bibinfo{person}{Haikun Huang}, {and} \bibinfo{person}{Lap-Fai Yu}.} \bibinfo{year}{2022}\natexlab{}.
\newblock \showarticletitle{{WFH}-{VR}: {Teleoperating} a {Robot} {Arm} to Set a {Dining} {Table} Across the {Globe} via {Virtual} {Reality}}. In \bibinfo{booktitle}{\emph{2022 {IEEE}/{RSJ} {International} {Conference} on {Intelligent} {Robots} and {Systems} ({IROS})}}. \bibinfo{publisher}{IEEE}, \bibinfo{address}{Kyoto, Japan}, \bibinfo{pages}{4927--4934}.
\newblock
\showISBNx{978-1-66547-927-1}
\urldef\tempurl%
\url{https://doi.org/10.1109/IROS47612.2022.9981729}
\showDOI{\tempurl}


\bibitem[Zhang et~al\mbox{.}(2018)]%
        {zhang_deep_2018}
\bibfield{author}{\bibinfo{person}{Tianhao Zhang}, \bibinfo{person}{Zoe McCarthy}, \bibinfo{person}{Owen Jow}, \bibinfo{person}{Dennis Lee}, \bibinfo{person}{Xi Chen}, \bibinfo{person}{Ken Goldberg}, {and} \bibinfo{person}{Pieter Abbeel}.} \bibinfo{year}{2018}\natexlab{}.
\newblock \showarticletitle{{Deep {Imitation} {Learning} for {Complex} {Manipulation} {Tasks} From {Virtual} {Reality} {Teleoperation}}}. In \bibinfo{booktitle}{\emph{2018 {IEEE} {International} {Conference} on {Robotics} and {Automation} ({ICRA})}}. \bibinfo{pages}{5628--5635}.
\newblock
\urldef\tempurl%
\url{https://doi.org/10.1109/ICRA.2018.8461249}
\showDOI{\tempurl}
\newblock
\shownote{ISSN: 2577-087X}.


\end{thebibliography}
\newpage
\appendix
\section{User Study Demographics}

\begin{table}[!htbp]
    \centering
    \tiny
    \scalebox{0.99}{
    \begin{tabular}{p{0.5cm} p{0.5cm} p{1cm} p{1.7cm} p{3.5cm} p{4cm} p{1.7cm} p{1.0cm}}

    \toprule
    \textbf{ID} & \textbf{Age} & \textbf{Gender} & \textbf{Education Level} & \textbf{Major} & \textbf{Robot Programming Experience} & \textbf{AR/VR Experience} & \textbf{Handedness} \\

    \\\midrule

P1 & 26 & Female & Doctoral & Electrical and Computer Engineering & No & Yes, Quest 2, once & Right \\
\\\midrule
P2 & 21 & Female & Bachelor & CS and Physics & No& Yes, Oculus Rift, once & Right \\
\\\midrule
P3 & 21 & Male & Bachelor & Computer Engineering & No & Yes, Quest 2, monthly & Right \\
\\\midrule
P4 & 22 & Female & Bachelor & Psychology & No & No & Right \\
\\\midrule
P5 & 24 & Female & Master & Electrical Engineering & No & No& Right \\\midrule
P6 & 21 & Male & Bachelor & Computer engineering & Yes, Lego Mindstorms, yearly & Yes, Quest, yearly & Left \\\midrule
P7 & 23 & Female & Master & Electrical and computer engineering & No & Yes, Quest 2, Vision Pro, around 10 times in total & Right \\\midrule
P8 & 28 & Male & Doctoral & Materials Science and Engineering & No & Yes, HMD, weekly & Right \\\midrule
P9 & 23 & Female & Master & Materials science and engineering & No & No & Right \\\midrule
P10 & 26 & Male & Doctoral & Student & No & No & Right \\\midrule
P11 & 22 & Female & Bachelor & Japanese Major & No & No & Right \\\midrule
P12 & 25 & Male & Bachelor & Mechanical Engineering & Yes, Aubo, Robotis Servos, Autonomous Rovers, daily & Yes, Quest 2, once & Right \\\midrule
P13 & 29 & Male & Doctoral & Electrical engineering & No & Yes, Quest Pro, yearly & Right \\\midrule
P14 & 21 & Male & Bachelor & Computer Science & Yes, The tinkerkit Braccio using Arduino, weekly & Yes, Quest 2\&3, and the Oculus Rift, monthly & Right \\\midrule
P15 & 23 & Female & Bachelor & Law School & No  & Yes, Quest 2, once & Right \\\midrule
P16 & 26 & Female & Master & Electrical and Computer Engineering & No & No & Right \\\midrule
P17 & 22 & Male & Bachelor & Biomedical Science & No & No & Right \\\midrule
P18 & 23 & Male & Bachelor & Computer Science & Yes, small Arduino based rovers, once & No & Right \\
\bottomrule
    \end{tabular}}
    \caption{Demographics and prior experience in robot programming and XR.}
    \Description{
    This table displays the dographics and prior experience of the 18 users in our final study. The header from left to right describes the participant No., age, gender, education level, major, robot programming experience, XR experience, and handedness.
    }
    \label{tab:demographics}
\end{table}

\end{document}